\documentclass{article} 
\usepackage{collas2022_conference,times}

\usepackage{amsmath,amsfonts,bm}

\def\eqref#1{equation~\ref{#1}}

\def\1{\bm{1}}

\DeclareMathAlphabet{\mathsfit}{\encodingdefault}{\sfdefault}{m}{sl}
\SetMathAlphabet{\mathsfit}{bold}{\encodingdefault}{\sfdefault}{bx}{n}

\DeclareMathOperator*{\argmax}{arg\,max}

\usepackage{hyperref}
\hypersetup{
    colorlinks=true,
    linkcolor=red,
    filecolor=magenta,      
    urlcolor=blue,
    citecolor=purple,
    pdftitle={Probing Transfer in Deep Reinforcement Learning without Task Engineering},
    pdfpagemode=FullScreen,
    }
    
\usepackage{graphicx}
\usepackage{array}
\usepackage{multirow}
\usepackage{multicol}
\usepackage{adjustbox}
\usepackage[inline]{enumitem}
\usepackage[symbol]{footmisc}
\usepackage{siunitx}
\usepackage{verbatim}

\makeatletter
\newcommand\footnoteref[1]{\protected@xdef\@thefnmark{\ref{#1}}\@footnotemark}
\makeatother

\newcolumntype{C}[1]{>{\centering\arraybackslash}m{#1}}
\newcolumntype{L}[1]{>{\raggedright\arraybackslash}m{#1}}

\newlist{inparaenum}{enumerate*}{2}
\setlist[inparaenum]{nosep}
\setlist[inparaenum,1]{label=(\bfseries\arabic*),itemjoin=.\ }
\setlist[inparaenum,2]{label=(\bfseries\arabic{inparaenumi}-\alph*),itemjoin=;\ }

\newlist{inparaenumquestions}{enumerate*}{2}
\setlist[inparaenumquestions]{nosep}
\setlist[inparaenumquestions,1]{label=(\bfseries\arabic*),itemjoin=\ }
\setlist[inparaenumquestions,2]{label=(\bfseries\arabic{inparaenumquestionsi}-\alph*),itemjoin=;\ }

\title{Probing Transfer in Deep Reinforcement Learning without Task Engineering}

\author{Andrei A.~Rusu, Sebastian Flennerhag, Dushyant Rao, Razvan Pascanu, Raia Hadsell  \\
DeepMind, UK \\
\texttt{\{andrei, flennerhag, dushyantr, razp, raia\}@deepmind.com} \\
}

\collasfinalcopy 

\begin{document}

\maketitle

\begin{abstract}
We evaluate the use of original game curricula supported by the Atari 2600 console as a heterogeneous transfer benchmark for deep reinforcement learning agents. Game designers created curricula using combinations of several discrete modifications to the basic versions of games such as Space Invaders, Breakout and Freeway, making them progressively more challenging for human players. By formally organising these modifications into several factors of variation, we are able to show that Analyses of Variance (ANOVA) are a potent tool for studying the effects of human-relevant domain changes on the learning and transfer performance of a deep reinforcement learning agent.
Since no manual task engineering is needed on our part, leveraging the original multi-factorial design avoids the pitfalls of unintentionally biasing the experimental setup.
We find that game design factors have a large and statistically significant impact on an agent's ability to learn, and so do their combinatorial interactions. Furthermore, we show that zero-shot transfer from the basic games to their respective variations is possible, but the variance in performance is also largely explained by interactions between factors.
As such, we argue that Atari game curricula offer a challenging benchmark for transfer learning in RL, that can help the community better understand the generalisation capabilities of RL agents along dimensions which meaningfully impact human generalisation performance.
As a start, we report that value-function finetuning of regularly trained agents achieves positive transfer in a majority of cases, but significant headroom for algorithmic innovation remains. We conclude with the observation that selective transfer from multiple variants could further improve performance.
\end{abstract}

\section{Introduction}
A key open challenge in artificial intelligence is training reinforcement learning (RL) agents which generally achieve high returns when faced with critical changes to their environments \citep{Schaul2018barbados}, motivated by impressive flexibility of animal and human learning. One way to approach the problem is through the prism of \emph{generalisation} across related but distinct environments, also called \emph{transfer learning} \citep{Pan2009survey, Taylor2009transfer} and comprehensively reviewed in the RL setting by \citet{Zhu2020survey}.
Many purpose-built benchmarks serve investigations into more specific research questions, e.g. transfer learning in particular cases where additional assumptions hold. While useful for progress, the challenge of transfer learning in the more general case remains.

Motivated by visual observation similarity,
\citet{Machado2018revisiting} suggest using the newest iteration of the Atari Learning Environment (ALE) \citep{Bellemare13arcade}  to study the transfer
learning between single-player game variants, or ``flavours'', found in the curricula of many Atari game titles.
We will use the terms \emph{default} or \emph{basic} game interchangeably to refer to the environment
recommended by respective manuals as the entry point. We call all other distinct games \emph{variations} or \emph{variants} of their respective \emph{default} game.
Hence, each Atari game title we consider provides a curriculum consisting of a \emph{default} and its \emph{variants}, all designed to teach and challenge human players in novel ways.

Studying transfer \emph{within curricula} ensures that environments are related, and that meaningful knowledge reuse should be possible and beneficial.
Interestingly, differences in game variant dynamics, subtle changes in observations which are crucial for optimal behaviour, novel environment states, as well as new player abilities challenge \emph{unified} approaches to transfer learning across variations. \cite{Farebrother2018generalization} argue that Deep Q-Network (DQN) agents \citep{Mnih2015dqn}, trained with appropriate regularisation, can be effective in zero-shot and finetuning transfer scenarios. In this work we study the learning and transfer performance of an updated version of the DQN agent, called Rainbow-IQN \citep{Toromanoff2019rainbowiqn}.
We use ANOVA to quantitatively confirm the suspected link between game design factors and agent performance. While current approaches occasionally achieve meaningful transfer from default games, they have limited success for variations with several modifications. Our analyses reveal this is due to strong interaction effects between factors, not just the isolated effects of the modifications they introduce.
This reinforces the case for agents leveraging these systematic curricula, originally designed for human players, through transfer learning.

{\bf Contributions:} \begin{inparaenum}
    \item We show that discrete changes to Atari game environments, challenging for human players, also modulate the performance of a popular model-free deep reinforcement learning algorithm starting from scratch
    \item Zero-shot transfer of policies trained on one game variation and tested on others can be significant, but performance is far from uniform across respective environments
    \item Interestingly, zero-shot transfer variance from default game experts is also well explained by game design factors, especially their interactions
    \item We empirically evaluate the performance of value-function finetuning, a general transfer learning technique compatible with model-free deep RL, and confirm that it can lead to positive transfer from basic game experts 
    \item We point out that more complex challenges of transfer with deep RL are captured by Atari game variations, e.g. appropriate source task selection, fast policy transfer and evaluation using experts, as well as data efficient behaviour adaptation more generally
\end{inparaenum}.

\section{Background}
\label{section:transfer}

\newcommand{\mdp}{\mathcal{M}}
\newcommand{\states}{\mathcal{X}}
\newcommand{\actions}{\mathcal{A}}
\newcommand{\rewards}{\mathcal{R}}
\newcommand{\transitions}{\mathcal{T}}
\newcommand{\policy}{\pi}
\newcommand{\reals}{\mathbb{R}}
\newcommand{\probs}{\left[0, 1\right]}
\newcommand{\expectation}[2]{\mathbb{E}_{#1}\left[ #2 \right]}
\renewcommand{\argmax}[1]{\underset{#1}{\operatorname{arg}\,\operatorname{max}}\;}

{\bf Reinforcement Learning.} A Markov Decision Processes (MDP) \citep{Puterman1994markov} is the classic abstraction used to characterise  the sequential interaction loop between an action taking agent and its environment, which responds with observations and rewards \citep{Sutton2018reinforcement}. Formally, a finite MDP is a tuple $\mdp = \langle\states, \actions, \transitions, \rewards, \gamma \rangle$, where $\states$ is the state space, $\actions$ is the action space, both finite sets, $\transitions : \states \times \actions \times \states \mapsto \probs$ is the stochastic transition function which maps each state and action to a probability distribution over possible future states $\transitions (x, a, x') = P(x'| x, a)$, $\rewards : \states \times \actions \times \states \mapsto \reals$ is the reward distribution function, with $r : \states \times \actions \mapsto \reals$ the expected immediate reward $r(x, a) = \expectation{\transitions}{\mathcal{R} (x, a, x') }$, and $\gamma \in [0, 1]$ is the discount factor \citep{Bellman1957markovian}. A stochastic policy function is the action selection strategy $\policy : \states \times \actions \mapsto \probs$ which maps states to probability distributions over actions $\policy(x) = P(a | x)$. The discounted sum of future rewards is the random variable $Z_{\mdp}^{\policy} (x, a) = \sum_{t = 0}^{\infty}  \gamma^t  r(x_t, a_t)$, where $x_0 = x$, $a_0 = a$, $x_t \sim \transitions(\cdot|x_{t-1}, a_{t-1})$ and $a_t \sim \policy(\cdot|x_t)$. Given an MDP $\mdp$ and a policy $\policy$, the value function is the expectation over the discounted sum of future rewards, also called the expected return: $V_{\mdp}^{\policy} (x) = \expectation{}{Z_{\mdp}^{\policy}(x, \policy(x))}$. The goal of Reinforcement Learning (RL) is  to find a  policy $\policy_{\mdp}^* : \states \mapsto \actions$ which is optimal, in the sense that it maximises expected return in $\mdp$. The state-action Q-function is defined as $Q_{\mdp}^{\policy} (x, a) = \expectation{}{ Z_{\mdp}^{\policy} (x, a)  }$ and it satisfies the Bellman equation $Q_{\mdp}^{\policy} (x, a) = \expectation{\transitions}{\rewards (x, a, x') } + \gamma \expectation{\transitions, \policy}{Q_{\mdp}^{\policy} (x', a')}$ for all states $x \in \states$ and actions $a \in \actions$. \citet{Mnih2013playing} adapted a reinforcement learning algorithm called Q-Learning \citep{Watkins1989learning} to train deep neural networks end-to-end, mastering several Atari 2600 games, with inputs consisting of high-dimensional observations, in the form of console screen pixels, and differences in game scores. For more details please consult Appendix~\ref{appendix:extended_background}.

Note that value functions depend critically on all aspects of the MDP. For any policy $\policy$, in general $Q_{\mdp}^{\policy} \ne Q_{\mdp'}^{\policy}$ for MDPs $\mdp$ and $\mdp'$ defined over the same state and action sets, with $\transitions \ne \transitions'$ or $\rewards \ne \rewards'$. Even if differences in dynamics or rewards are isolated to a subset of $\states \times \actions$, changes may be induced across the support of $Q_{\mdp'}^{\policy}$, since value functions are expectations over sums of discounted future rewards, issued according to $\rewards'$, along sequences of states decided entirely by the new environment dynamics $\transitions'$ when following a fixed behaviour policy $\policy$. Nevertheless, many particular cases of interest exist, which we discuss below.

{\bf Transfer learning.} Described and motivated in its general form by \citet{Caruana1997multitask, Thrun1998l2l, Bengio12transfer}, the goal of transfer learning is to use knowledge acquired from one or more \emph{source} tasks to improve the learning process, or its outcomes, for one or more \emph{target} tasks, e.g. by using fewer resources compared to learning from scratch. When this is achieved, we call it \emph{positive} transfer.
We often further qualify transfer learning by the metric used to measure specific effects. Several transfer learning metrics have been defined for the RL setting \citep{Taylor2009transfer}, but none capture all aspects of interest on their own. A first metric we use is ``jumpstart'' or ``zero-shot'' transfer, which is performance on the target task before gaining access to its data. 
Another highly relevant metric is ``performance with fixed training epochs'' \citep{Zhu2020survey}, defined as returns achieved in the target task after using fixed computation and data budgets under transfer learning conditions.

One way to classify such approaches in RL is by the format of knowledge being transferred \citep{Zhu2020survey}, commonly: datasets, predictions, and/or parameters of neural networks encoding representations of observations, policies, value-functions or approximate state-transition ``world models'' acquired using source tasks. Another way to classify transfer learning approaches in RL is by their respective sets of \emph{assumptions} about source and target tasks, also well illustrated by their associated benchmark domains: \begin{inparaenum}
\item Differences are limited to observations, but the underlying MDP is the same, e.g. domain adaptation and randomisation \citep{Tobin2017random}, mastered through the generalisation abilities of large models \citep{Cobbe2019quantifying, Cobbe2020leveraging}
\item MDP states and dynamics are the same, but reward functions are different: successor features and representations \citep{Barreto2017successor}
\item Overlapping state spaces with similar dynamics, e.g. multitask training and policy transfer \citep{Rusu2016distillation, schmitt2018kickstarting}, contextual parametric approaches, e.g. UVFAs \citep{Schaul15uva} and collections of skills/goals/options \citep{Barreto2019option}
\item Suspected but unqualified overlaps between tasks \citep{Parisotto2016actor, Rusu2016progressive, Rusu2017sim2real}
\item Large, curated collections of similar environments designed to facilitate complex transfer and fast adaptation through meta-learning \citep{Yu2020metaworld, Hospedales2020meta} \end{inparaenum}. All these works capture important sub-problems of interest, and clever design of specialised benchmarks has greatly aided progress.
Atari game variations \citep{Machado2018revisiting} offer the exciting prospect of direct comparisons between generalisations of transfer learning methods originally developed under different sets of assumptions, by measuring their performance along dimensions of variation which are meaningful and challenging for human players, one of many interesting criteria cutting across specialised paradigms.

\begin{figure}[h]
\begin{center}
\includegraphics[width=0.96\textwidth]{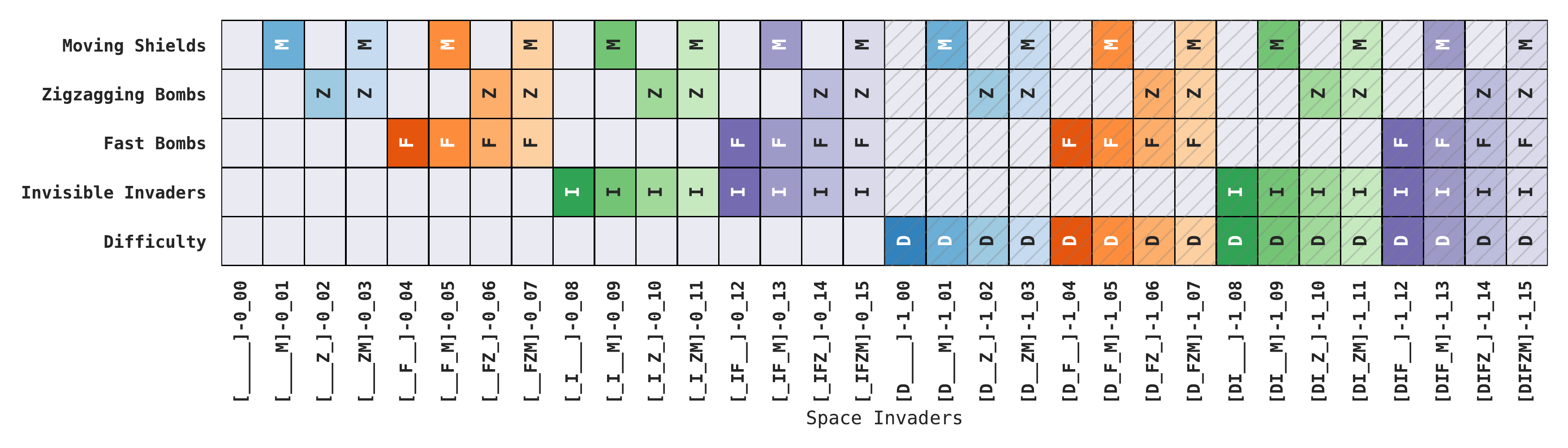}
\includegraphics[width=0.59\textwidth]{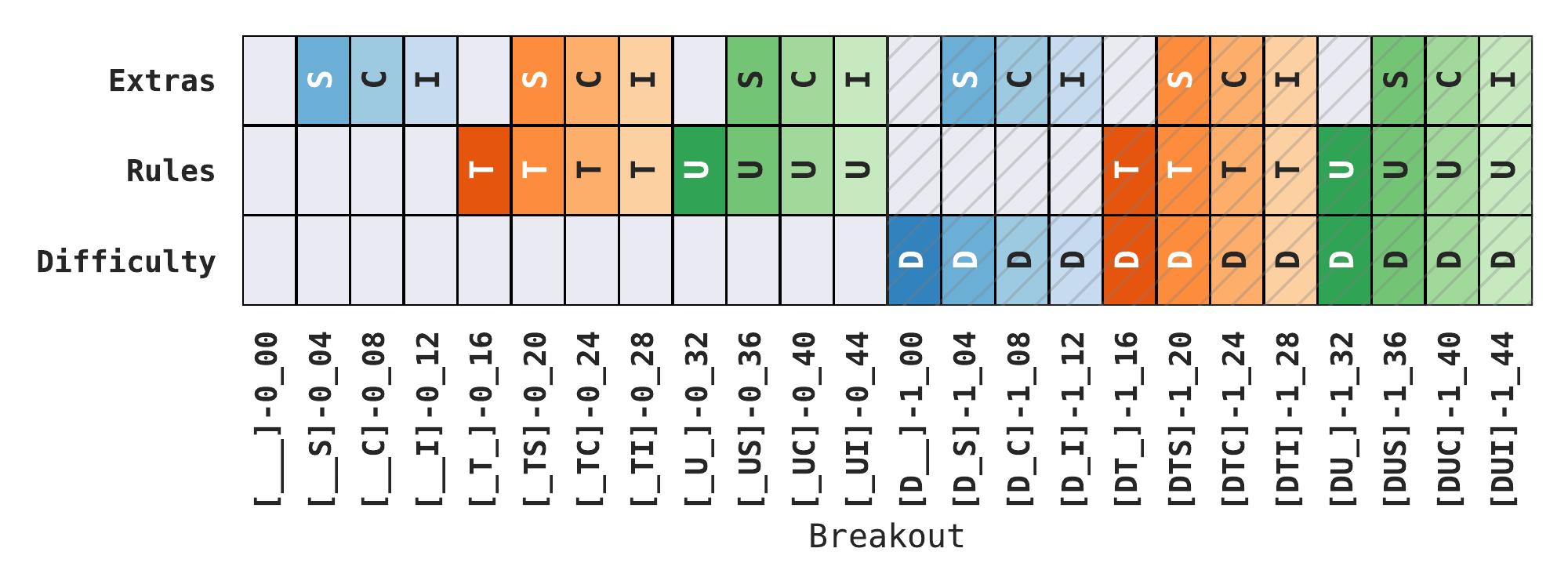}
\includegraphics[width=0.38\textwidth]{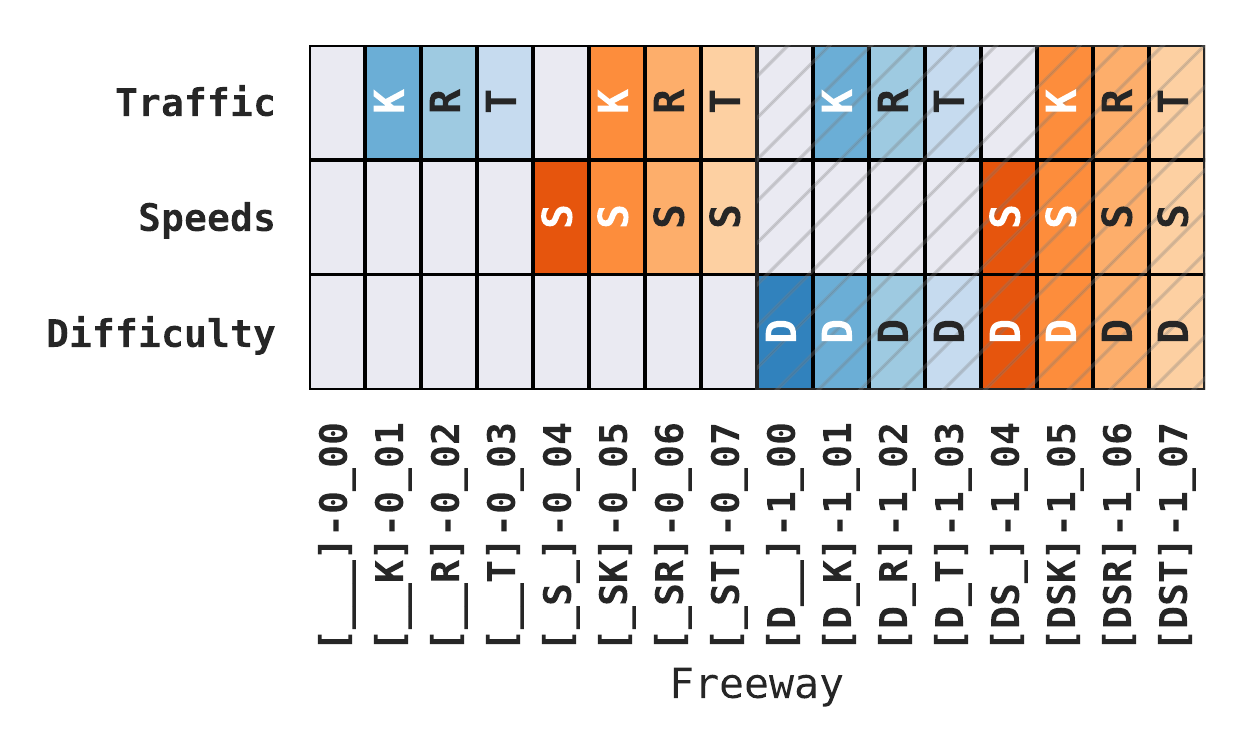}
\end{center}
\caption{Variant naming convention and factorial design matrices for Space Invaders (top), Breakout (bottom left) and Freeway (bottom right). All factors of variation are categorical and highlighted only when taking different values from default games. Binary factors are abbreviated by their initial, e.g. ``Moving Shields'' is plotted as `M'. Non-binary factors are: Breakout ``Rules'' with additional values: `T' for ``Timed Breakout'', or `U' for ``Breakthru''; Breakout ``Extras'' with additional values: `S' for ``Steerable'', `C' for ``Catch'', or `I' for ``Invisible''; Freeway ``Traffic'' with levels: `K' for ``Thick'', `R' for ``Thicker'', or `T' for ``Thickest''.
Colours indicate the ``game mode'', and hatching denotes the ``difficulty'' switch being activated. Colours and horizontal labels correspond to those used in subsequent figures. Labels include factor abbreviations to supplement the ALE naming conventions. \label{figures:variant_legend}}
\end{figure}

\subsection{ALE Atari as a Transfer Learning Benchmark}

With the first version of the Arcade Learning Environment (ALE), \citet{Bellemare13arcade} gave access to over $50$ different Atari $2600$ game titles through a unified observation and action interface, modelled after the standard reinforcement learning loop. The ALE proved an excellent development test-bed for building general deep RL agents, in no small part due to its diversity and being devoid of experimenter bias, since games were not modified by researchers.

The second and latest version of the ALE \citep{Machado2018revisiting} opens up a wealth of game variations for transfer learning research. This is achieved by emulating functions of ``difficulty'' and ``game select'' switches for single-player games. The original cartridges of many game titles came with variations which could be selected and played using these switches.
We assign unique identifiers of game variants using the $\mathrm{X}\_\mathrm{YZ}$ notation, with $\mathrm{X} \in \{0, 1\}$ denoting the position of the ``difficulty'' switch, and $\mathrm{YZ} \in \{00, 01, 02, \dots\}$ indicating the selected ``game mode''. Value ranges are game specific and identical to those used by the ALE code-base \citep{Machado2018revisiting}. For example, the entry game version of each title is denoted as $0\_00$, which we call the default or basic game.
All game variations are different environments that still feature the main concepts of the default. Furthermore, the variants were designed to serve as curricula for human players, hence positive transfer should be possible. Most importantly for our purposes, the ALE remains largely free of experimenter bias. \citet{Farebrother2018generalization} selected a few representative variants from four game titles for experiments. We aim to analyse entire curricula, thus we consider the top three most popular titles from their list, according to sales, and use all their variations, for a total of $72$ distinct environments.

The original designers created game variations using combinations of discrete modifications to a default ``entry'' game,  qualitatively described in accompanying game manuals. We organise these discrete modification into formal factors of variation, following closely the original game design matrices sometimes explicitly plotted in manuals. In Figure~\ref{figures:variant_legend} we formalise variant naming conventions relative to default games. We briefly explain the meanings of all design factors below, to help build an intuitive understanding of this heterogeneous collection of transfer learning scenarios.

{\bf Space Invaders.} The player controls an Earth-based laser cannon which can move horizontally along the bottom of the screen. There is a grid of approaching aliens above the player, shooting laser bombs, and the objective of the game is to eliminate all the aliens before they reach the Earth, especially their Command Ships. Three destructible shields above the cannon offer some protection. The game ends once any alien reaches the Earth or the player’s cannon is hit a third time with laser bombs. Game variants are combinations of five binary factors which, when activated, modify the default:
\begin{inparaenum}
    \item The difficulty switch widens the player’s laser cannon, making it an easier target for enemy laser bombs
    \item Shields move, and thus are harder to use
    \item Enemy laser bombs zigzag, which makes them harder to predict
    \item Enemy laser bombs drop faster
    \item Invaders are invisible, and only hitting one with the laser briefly illuminates the others
\end{inparaenum}. This creates a total of $32$ variants of Space Invader.

{\bf Breakout.} In order to achieve the best score in any variant, the players need to completely break walls with six layers of coloured bricks by knocking off said bricks with a ball, served and bounced towards the wall with a controllable paddle, which they can move horizontally along the bottom of the screen. The ball will also bounce off of screen edges, except for the bottom edge, where balls either come in contact with the paddle or are lost. Players have a total of five balls at their disposal, and the game ends when all balls are lost. Points are scored when bricks are knocked off the wall, and the ball accelerates on contact with bricks in the top three rows or after twelve consecutive hits. Game variants are created by all combinations of three factors:
\begin{inparaenum}
    \item The binary difficulty switch reduces the paddle's width by a quarter, making it easier to miss the ball
    \item The precise rules which determine how points accumulate:
        \begin{inparaenum}
            \item with standard ``Breakout'' rules, e.g. in the default game $(0\_00)$, the player must completely break down two walls of bricks, one after the other, while loosing the fewest balls
            \item under ``Timed Breakout'' rules, the player must completely break a single wall as fast as possible, no matter how many of the five balls are used
            \item with ``Breakthru'' rules, the player needs to break two walls consecutively, but the ball does not bounce off bricks, unlike in previous variants; the ball keeps going through the wall, quickly picking up speed and accumulating points 
        \end{inparaenum}
    \item How the ball is aimed at the wall and ``extras'':
        \begin{inparaenum}
            \item the ball simply bounces off the paddle at a position dependent angle
            \item the player can also steer the ball in flight
            \item the player is also able catch the ball and release it at a slower speed
            \item the wall is invisible and is only briefly illuminated when a brick if knocked off, so the player needs to remember its configuration in order to aim effectively
        \end{inparaenum}
\end{inparaenum}.
All factors together define $24$ variants of Breakout.

{\bf Freeway.} In all variants, the goal of the player is to safely get a chicken across ten lanes of freeway traffic as many times as possible in 2 minutes and $16$ seconds. Variants are constructed as combinations of three design factors:
\begin{inparaenum}
    \item The difficulty switch controls whether the chicken is knocked back one lane, or all the way to the kerb, when hit by incoming vehicles
    \item Traffic ``thickness'', defined as four levels of traffic density
    \item Vehicle speeds across lanes, either constant or randomised
\end{inparaenum}.  Hence, there are $16$ variants of Freeway.

\section{Methodology}

{\bf Experimental setup.} 
Following the latest ALE benchmark
recommendations \citep{Machado2018revisiting}, information about player ``lives'' is not disclosed to agents, and stochasticity is introduced using randomised action replay with probability $25\%$, also known as ``sticky actions''.
Irrespective of redundancies, agents act using the full set of $18$ actions.
Environment observations (Atari screen frames) were pre-processed according to standard practice \citep{Mnih2015dqn, Hessel2018rainbow}.
We report agent returns at the end of training, averaged over the last 2 million steps.

{\bf Expert Agent Training.} We use the Rainbow-IQN model-free deep reinforcement learning algorithm \citep{Hessel2018rainbow, Dabney2018iqn, Toromanoff2019rainbowiqn} since it is available to the community in several implementations, e.g. \citet{Castro2018dopamine}, and is effective with widely available commodity hardware and open-source software.

Rainbow \citep{Hessel2018rainbow} collects a number of improvements to DQN \citep{Mnih2013playing, Mnih2015dqn}, of which we used Double Q-Learning \citep{Van2016deep}, Prioritised Replay \citep{Schaul2016prioritized} and multi-step learning \citep{Sutton2018reinforcement}.  Following \citet{Castro2018dopamine}, we did not use the dueling network architecture \citep{Wang2016sample} or noisy networks \citep{Fortunato2018noisy}. We  replaced the distributional RL approach C51 \citep{Bellemare2017distributional} with its more general form (IQN), since it has been shown to be superior \citep{Dabney2018iqn, Toromanoff2019rainbowiqn}.

We used the standard limit on environment interactions of $200$ million steps \citep{Mnih2015dqn, Mnih2016asynchronous, Dabney2018iqn, Machado2018revisiting, Toromanoff2019rainbowiqn}. Our study is the first to report Rainbow-IQN results for Atari game variants, hence we use independent hyper-parameter selection for expert training and finetuning experiments.

{\bf Expert Agent Finetuning.} The experimental setup for finetuning follows closely that for agent training, except for two important differences: \begin{inparaenumquestions}
    \item We used $10$ million environment steps to adapt to new variants, following \cite{Farebrother2018generalization}, which is $20\times$ less data than what variant-experts are trained with. The aim of transfer learning is to reduce resources needed for acquiring new knowledge and behaviours, hence we are interested in improving sample complexity using transfer.
    \item The hyper-parameter grid was slightly adapted in order to improve chances of fast learning in this reduced data regime. Further details, including parameter grids, are listed in Appendix~\ref{appendix:agent_training_details}.
\end{inparaenumquestions}

{\bf Statistical Analyses.} A multi-factor Analyses of Variance (ANOVA) \citep{girden1992anova} is a statistical procedure which investigates the influence that two or more independent variables, or factors, have on a dependent variable. The factors can take two or more categorical values, and experimental designs are often balanced: they use equal numbers of independent observations for all combinations of factor values, no less than three. Other assumptions are normality of deviations from group means, and equal or similar variances, see the discussion in Appendix~\ref{appendix:statistical_analyses}. ANOVA can be used to reveal statistically significant evidence against the \emph{null-hypothesis} that group means are all the same. We study the influence of game design factors, discrete modification to default games, on average returns of learned policies.

{\bf Study Limitations.} The relationship between hyper-parameters and quality of policies learned with deep RL is complex and poorly understood. \citet{Mnih2015dqn} used task-agnostic DQN hyper-parameters, tuned across a few basic games. Later works introduce several interacting modification, including changes to sets of hyper-parameters \citep{Schaul2016prioritized, Dabney2018iqn, Toromanoff2019rainbowiqn}. With the risk of maximisation bias, we used grid searches per game variant in order to mitigate the greater risks of incorrect inferences about Atari curricula due to poor hyper-parameter settings or divergence. Future works may perform fine-grained sensitivity analyses, using more than two random seeds. Due to computational limitations, we instead report the returns of the top-three agents from our grids.

\section{Experiments}

We would like to characterise the performance of general transfer learning strategies across heterogeneous scenarios,
designed to progressively challenge human players.
While the shapes of learning curves are expected to be different between game variants, agent performance is thought to be bounded only by its inductive biases, the learning algorithm and practical limitation on resources, such as computation or interactions \citep{Silver2021enough}.
We say that some game variant is ``harder'' for a given agent if the average returns of learned policies are lower given equal resource budgets.
We aim to explain this in terms of the factors of variation which combinatorially define environments within curricula, knowing that they also impact human player performance.
Comparing raw scores across the benchmark is difficult if the underlying tasks are ``harder'' to learn from scratch, or if variants have different scoring scales, as is sometimes the case here. Hence, we must first establish the performance of our agent learning game variants in isolation. On this basis, we then introduce a relative scoring scheme which enables conclusions to be drawn at the benchmark level.

\begin{figure}[h]
\begin{center}
\includegraphics[width=0.99\textwidth]{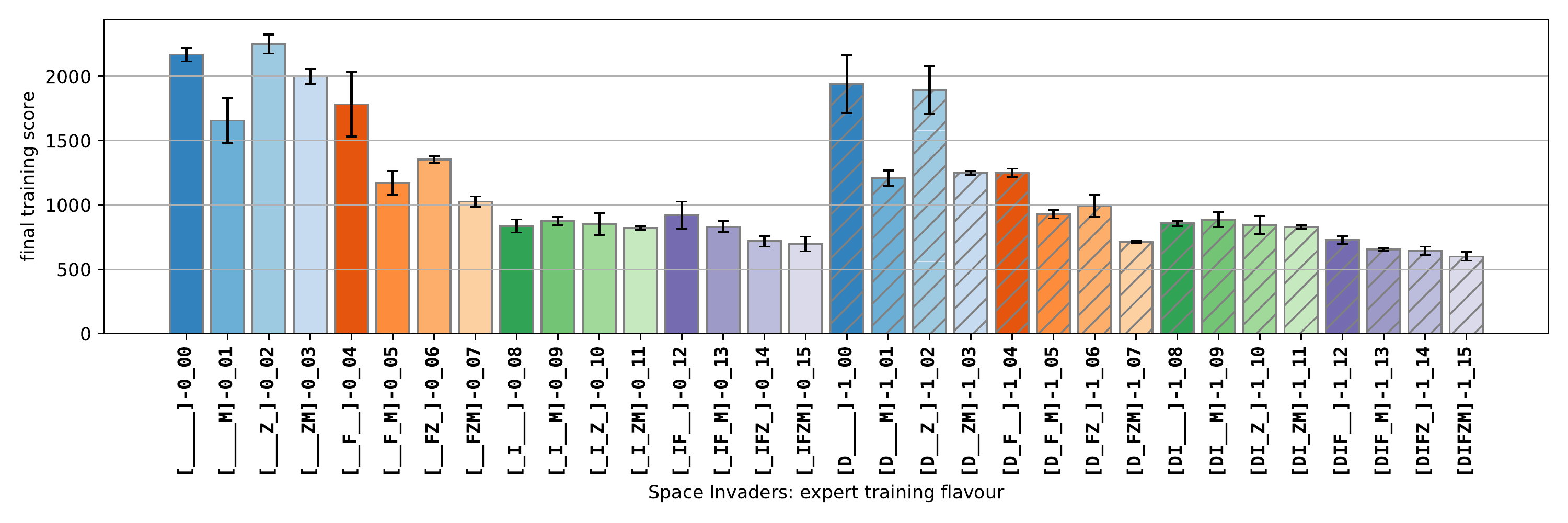}
\includegraphics[width=0.61\textwidth]{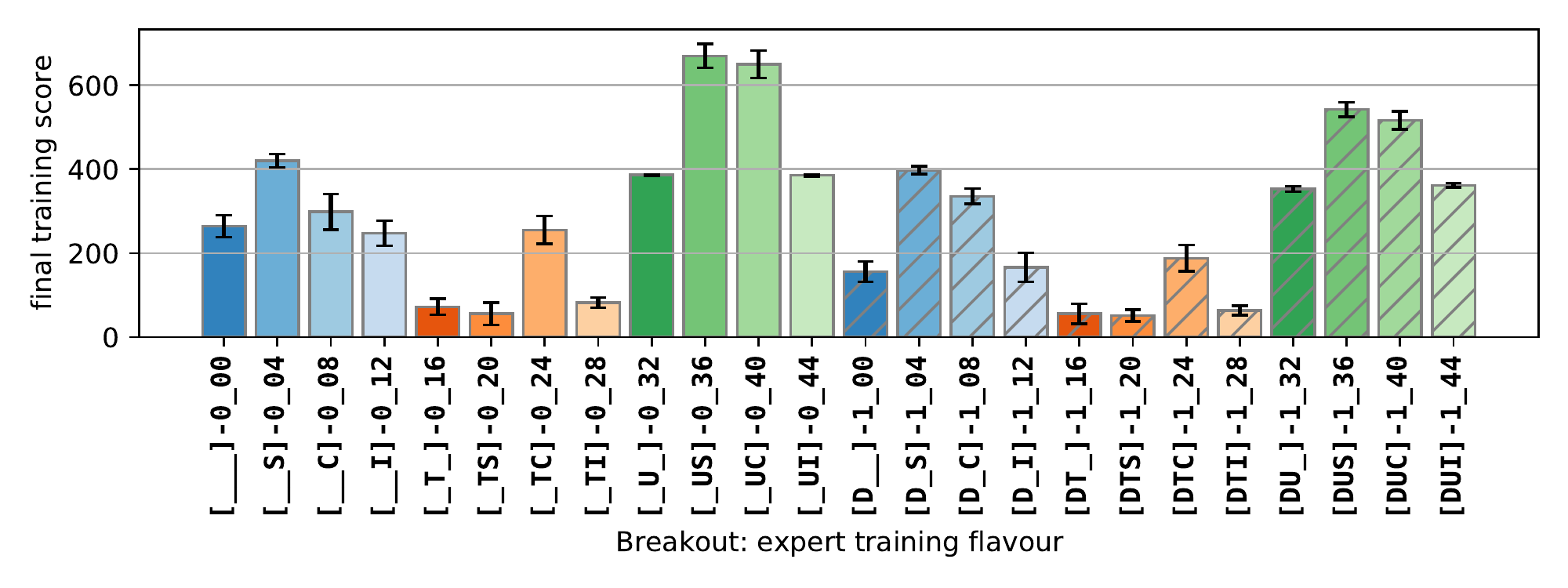}
\includegraphics[width=0.38\textwidth]{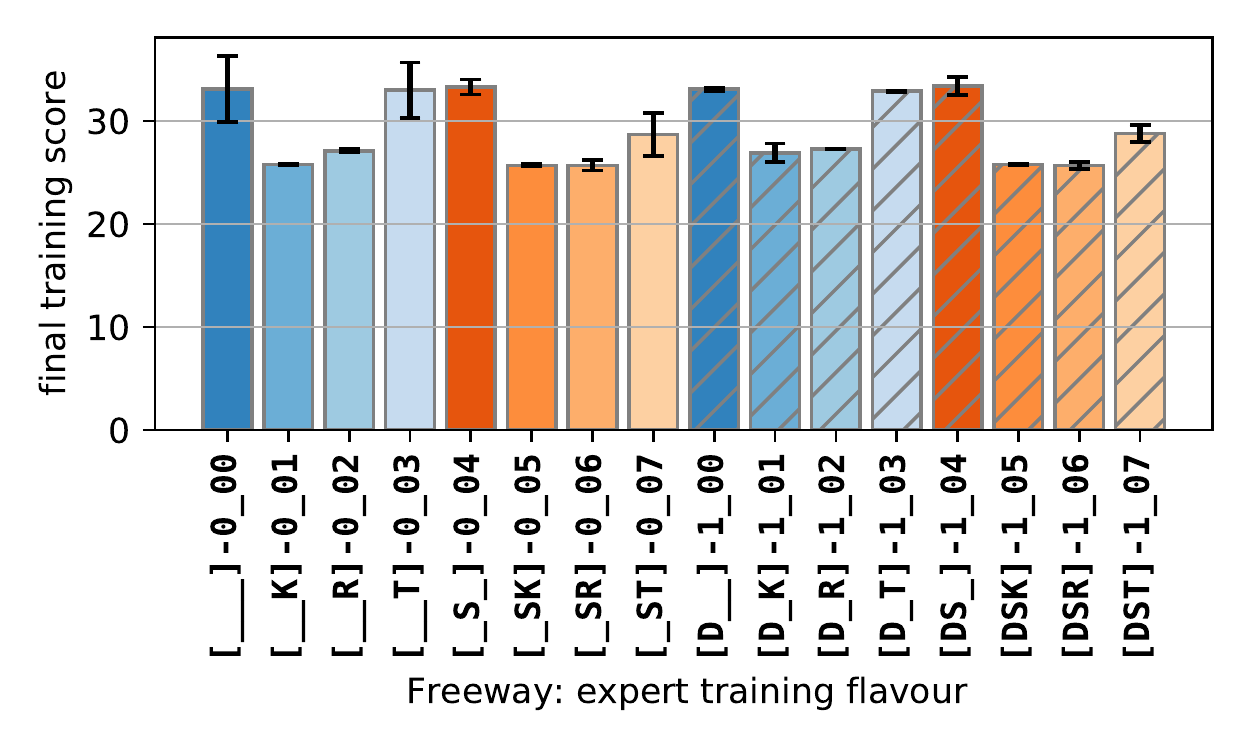}
\end{center}
\caption{Variant-Expert final raw score distributions, means and standard deviations across the top three experts in their respective hyper-parameter grids, reported for Space Invaders (top), Breakout (bottom left) and Freeway variants (bottom right).  In all plots, hatching denotes the ``difficulty'' switch being activated, and colours indicate the ``game mode''. See factorial design matrices for meanings of colour groups (Figure~\ref{figures:variant_legend}).\label{figures:experts}}
\end{figure}

\subsection{Training Variant-Experts}

We aim to answer the following questions: \begin{inparaenumquestions}
    \item Does our agent achieve similar levels of performance when learning variants of the same game from scratch?
    \item If not, what explains differences in performance across variants?
\end{inparaenumquestions}
In Figure~\ref{figures:experts} we report means and standard deviations of top three variant-experts trained from scratch for 200 million environment steps.
Our scores on default games are largely in line with those reported by other implementations of basic IQN agents, e.g. \cite{Castro2018dopamine}. On remaining variants, we find that Rainbow-IQN performance varies widely for some game titles, less so on Freeway, even with independent hyper-parameter tuning. Overall, our agent's scores are below variant performance ceilings, suggesting that some game variations may be harder to learn. While this may be at times due to the agent's inductive biases---in particular, invisible objects---it is unlikely to fully explain observed variation, because such changes to basic games do not always have a detrimental impact on their own. For example, experts achieve significantly different levels of performance on identical variants of Space Invaders, except for visible vs invisible aliens ($0\_00$ vs.$0\_08$). However, experts have virtually equal performance with the same change to environment observations across Breakout variants, visible vs invisible walls of bricks (e.g. $0\_00$ vs. $0\_12$).
Rather, we hypothesise that some variants are inherently harder to learn for the chosen agent, and thus a fixed training budget leads to performance differences across variations. This imposes a challenging bottleneck for transfer learning, which places high priority on limiting the resources expended for acquiring behaviours which maximise cumulative returns.

\begin{table}[h]
\caption{Multi-Factor ANOVA (type 3) of Expert Scores vs. design factors.}
\vspace{10pt}
\label{tables:anova_experts}
\begin{minipage}{\linewidth}
\resizebox{.35\linewidth}{!}{
\centering
\begin{tabular}{lcccc}
{\bf Space Invaders}        &   $df$        &   $F$             &   $PR(>F)$        \\ \hline
Difficulty		            &   $1.0$       &   $ 115.90$       &   $5.32e-16$\footnote{\label{mean_diff_significant}statistically significant after appropriate Bonferroni correction.}      \\
Invisible Invaders      	&   $1.0$       &   $1000.59$       &   $8.67e-41$\footnoteref{mean_diff_significant}      \\
Fast Bombs      		    &   $1.0$       &   $ 314.09$       &   $2.24e-26$\footnoteref{mean_diff_significant}      \\
Zigzagging Bombs		    &   $1.0$       &   $  12.08$       &   $9.22e-04$\footnoteref{mean_diff_significant}      \\
Moving Shields	            &   $1.0$       &   $ 125.51$       &   $9.93e-17$\footnoteref{mean_diff_significant}      \\ \hline
{\it Interaction} 		    &   $26.0$      &   $1341.37$       &   $5.33e-78$\footnote{\label{ombibus_significant}statistically significant interaction effect at level: $\alpha = 0.05$.}      \\
\end{tabular}
}
\hfill
\resizebox{.32\linewidth}{!}{
\centering
\begin{tabular}{lcccc}
{\bf Breakout}          &   $df$        &   $F$             &   $PR(>F)$        \\ \hline
Difficulty      		&   $1$         &   $  56.28$       &   $1.26e-09$\footnoteref{mean_diff_significant}       \\
Rules       			&   $2$         &   $1368.66$       &   $4.71e-43$\footnoteref{mean_diff_significant}       \\
Extras      			&   $3$         &   $ 179.84$       &   $4.22e-26$\footnoteref{mean_diff_significant}       \\ \hline
{\it Interaction}      	&   $17$        &   $  23.04$       &   $1.91e-17$\footnoteref{ombibus_significant}         \\
\end{tabular}
}
\hfill
\resizebox{.32\linewidth}{!}{
\centering
\begin{tabular}{lcccc}
{\bf Freeway}           &   $df$        &   $F$             &   $PR(>F)$        \\ \hline
Difficulty		        &   $1$         &   $  1.39$        &   $2.46e-01$      \\
Speeds			        &   $1$         &   $ 91.38$        &   $6.75e-11$\footnoteref{mean_diff_significant}      \\
Traffic			        &   $3$         &   $453.84$        &   $2.73e-26$\footnoteref{mean_diff_significant}      \\ \hline
{\it Interaction} 		&   $10$        &   $274.76$        &   $4.42e-28$\footnoteref{ombibus_significant}      \\
\end{tabular}
}
\end{minipage}
\end{table}

{\bf Statistical Analyses.} We verify that variants have meaningful impact on learning by rejecting the null hypothesis that expert performance is the same across all combinations of the design factors which define game variants. We perform multi-factor Analysis of Variance (ANOVA) tests separately for each game title, and report results in Table~\ref{tables:anova_experts}. Interaction effects are statistically significant in all cases, supporting the hypothesis that---although conceptually similar---design factors introduce significant changes to agent learning dynamics, even within the same game. 
In the post-hoc analyses of differences between groups we found that all factors have a statistically significant effect on agent learning from scratch apart from the ``difficulty'' factor in Freeway. 
Further details can be found in Appendix~\ref{appendix:statistical_analyses}.

{\bf Discussion.} Some differences in agent performance across variants were expected due to the qualitative impact of some factors on human play, e.g. changes in the dynamics of state transitions between ``Breakout'' and ``Breakthru'' variants, changes in observations such as visible vs. invisible invaders, or the effect of the difficulty switch. It is somewhat surprising that agents learn Freeway variants to similar performance, irrespective of difficulty level. 
Generally, we find that combinations of factors seem to induce substantially different learning dynamics, each having a significant effect on the final performance of the agent, at least within the standard training budget considered here. Hence, Rainbow-IQN demonstrates different levels of data efficiency across variants, despite their many similarities. A good amount of the variation can be explained by game design factors, which further emphasises the value of this benchmark for systematic and unbiased investigations of transfer with reinforcement learning algorithms.

Considering these empirical findings, as well as the possibility of different asymptotic performance ceilings for different variants, the remainder of the paper reports results relative to variant-expert performance, in order to meaningfully compare transfer learning across variations and curricula on a unified scale. We stress that variant-expert agents are generally not near-optimal.
The most surprising results were the comparatively low agent score on ``Timed Breakout'' variants (orange bars in the bottom left plot of Figure~\ref{figures:experts}), which have identical transition dynamics to regular ``Breakout'', but observations differ in that time in seconds is displayed instead of current score. Overall, these findings imply that scores normalised by variant-expert performance can be higher than $100\%$. Nevertheless, a relative scoring scheme accounts for the main effects of factors of variation on our agent's data efficiency while learning particular game variants. This is essential for meaningful comparisons of transfer learning within and across game curricula.

\begin{figure}[h]
\begin{center}
\includegraphics[width=0.99\textwidth]{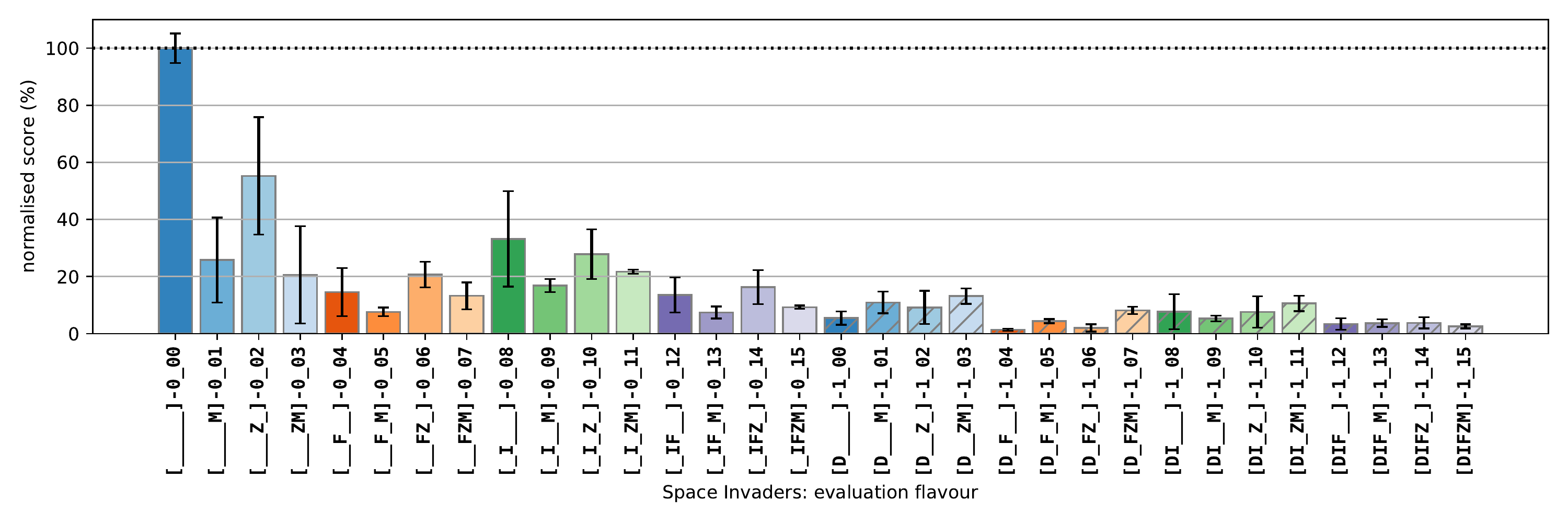}
\includegraphics[width=0.61\textwidth]{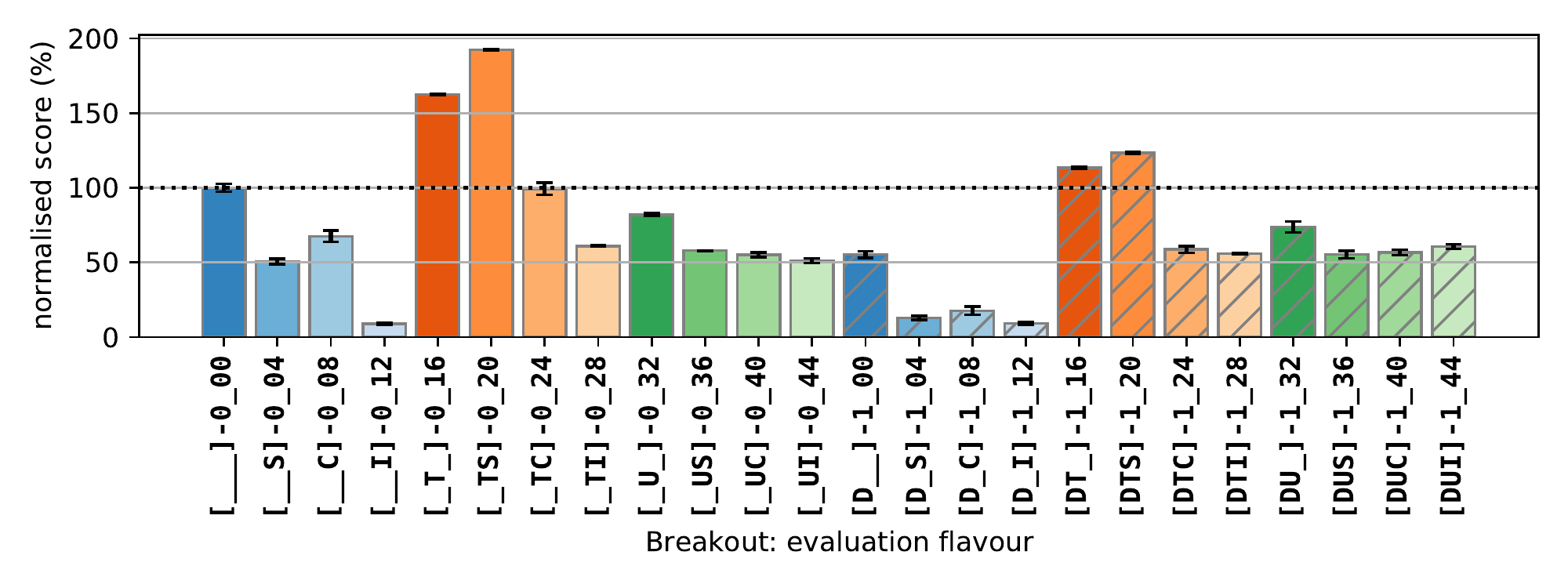}
\includegraphics[width=0.38\textwidth]{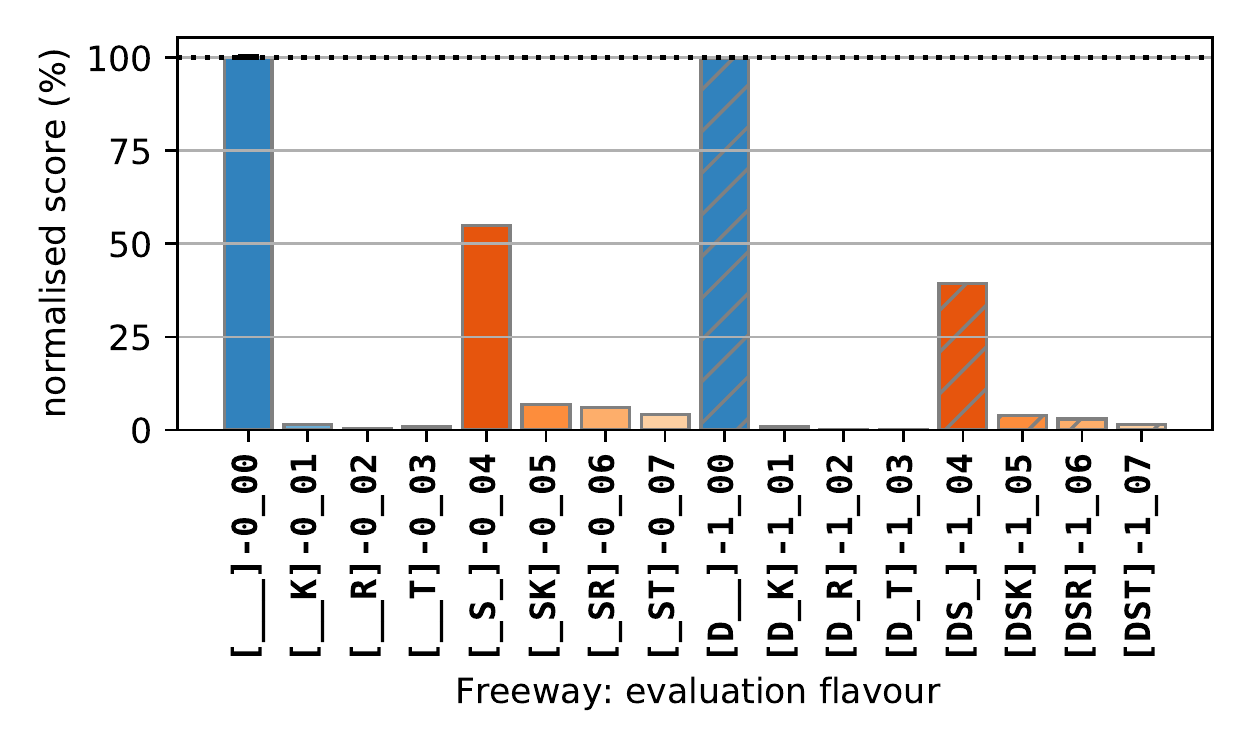}
\end{center}
\caption{Zero-shot transfer performance of default game experts $(0\_00)$ on all other variants, plotted as normalised scores (percentages) of the final performance of respective variant-experts, reported for Space Invaders (top), Breakout (bottom left) and Freeway variants (bottom right). Hatching denotes the ``difficulty'' switch being activated, and colours indicate the ``game mode''. Colour groups are meaningful with respect to the factorial design matrices of particular game variants  (Figure~\ref{figures:variant_legend}).\label{figures:default_jumpstart}}
\end{figure}

\begin{table}[h]
\caption{Multi-Factor ANOVA (type 3) of default game expert zero-shot transfer vs. design factors.}
\centering
\label{tables:anova_default}
\vspace{10pt}
\begin{minipage}{\linewidth}
\resizebox{.35\linewidth}{!}{
\centering
\begin{tabular}{lcccc}
{\bf Space Invaders}        &   $df$        &   $F$             &   $PR(>F)$        \\ \hline
Difficulty		            &   $1$         &   $319.42$        &   $1.43e-26$\footnote{\label{mean_diff_significant}statistically significant after appropriate Bonferroni correction.}      \\
Invisible Invaders	        &   $1$         &   $208.13$        &   $8.68e-22$\footnoteref{mean_diff_significant}      \\
Fast Bombs		            &   $1$         &   $250.95$        &   $7.92e-24$\footnoteref{mean_diff_significant}      \\
Zigzagging Bombs		    &   $1$         &   $  6.51$        &   $1.32e-02$      \\
Moving Shields	            &   $1$         &   $116.63$        &   $4.67e-16$\footnoteref{mean_diff_significant}      \\ \hline
{\it Interaction} 		    &   $26$        &   $117.61$        &   $1.57e-44$\footnote{\label{ombibus_significant}statistically significant interaction effect at level: $\alpha = 0.05$.}      \\
\end{tabular}
}
\hfill
\resizebox{.32\linewidth}{!}{
\centering
\begin{tabular}{lcccc}
{\bf Breakout}              &   $df$        &   $F$             &   $PR(>F)$        \\ \hline
Difficulty		            &   $1$         &   $182.22$        &   $5.83e-18$\footnoteref{mean_diff_significant}      \\
Rules			            &   $2$         &   $476.93$        &   $2.14e-32$\footnoteref{mean_diff_significant}      \\
Extras			            &   $3$         &   $177.63$        &   $5.53e-26$\footnoteref{mean_diff_significant}      \\ \hline
{\it Interaction} 	        &   $17$        &   $ 31.38$        &   $2.87e-20$\footnoteref{ombibus_significant}      \\
\end{tabular}
}
\hfill
\resizebox{.32\linewidth}{!}{
\centering
\begin{tabular}{lcccc}
{\bf Freeway}               &   $df$        &   $F$             &   $PR(>F)$        \\ \hline
Difficulty		            &   $1$         &   $  15.87$       &   $3.66e-04$\footnoteref{mean_diff_significant}      \\
Speeds			            &   $1$         &   $ 211.61$       &   $1.17e-15$\footnoteref{mean_diff_significant}      \\
Traffic			            &   $3$         &   $1524.22$       &   $1.36e-34$\footnoteref{mean_diff_significant}      \\ \hline
{\it Interaction} 		    &   $10$        &   $ 100.01$       &   $3.14e-21$\footnoteref{ombibus_significant}      \\
\end{tabular}
}
\end{minipage}
\end{table}

\subsection{Zero-shot Transfer using Default Game Experts}

Game variants go beyond defaults in ways which are interesting to humans, providing additional challenges by design.
Players will often become proficient in the default game, and then attempt to master others; indeed, several game manuals explicitly recommend this approach. Since deep neural networks are known to generalise to similar inputs, and since variants reuse the same visual elements (game sprites), we would like to answer the following questions: 
\begin{inparaenumquestions}
    \item How much transfer can be achieved purely through the generalisation power of deep neural networks trained on default games $(0\_00)$?
    \item Do factors of variation explain the success of zero-shot transfer \citep{Taylor2009transfer}?
\end{inparaenumquestions} 

Default game experts are evaluated, without further training, on all remaining variants of their respective games, and normalised scores are reported in Figure~\ref{figures:default_jumpstart}. Means and standard deviations of zero-shot transfer performance are plotted as percentages of the scores achieved by respective variant-experts, which were trained from scratch using 200 million interaction steps. Rainbow-IQN policies are derived using $\epsilon$-greedy action selection according to value function approximations output by deep neural networks. In the zero-shot transfer case, errors of such predictions are unbounded, but depend on the differences in reward functions and transition dynamics between the source and transfer target MDPs, as well as on the specific function approximation method used. In the case of deep neural networks we have no formal guarantees for the quality of prediction on environment states never encountered in the source task, especially for those which do not exist at all in the source. All things considered, a surprising amount of positive zero-shot transfer is observed in all game domains, due to the sheer generalisation power of deep neural networks. In some cases, e.g. ``Timed Breakout'' variants, zero-shot transfer results in superior policies to what variant-experts have learned, although data and computational budgets were identical. However, these are exceptions. In the majority of cases, learning on default games results in lower return policies on other variants. That said, the observed zero-shot transfer is not negligible, and it does not appear uniform across game variations, which we investigate next.

{\bf Statistical Analyses.} In order to answer the second question we check whether zero-shot performance is statistically indistinguishable across game variants, our null hypothesis of interest. We proceed with a multi-factor ANOVA of the dependent variable, here raw zero-shot transfer score, as a function of factors of variation in each game domain. Results reported in Table~\ref{tables:anova_default} indicate a statistically significant interactions in all games, meaning that factors play a role in the level of zero-shot transfer observed. In post-hoc analyses we must correct for multiple comparisons, and hence significance at level $\alpha = 0.05$ in Space Invaders is achieved for a corrected threshold of $\alpha_{c} = 0.0015$ or lower. We find statistically significant effects of all factors except the introduction of ``Zigzagging Bombs'' in Space Invaders.

{\bf Discussion.} Zero-shot transfer results provide evidence that the benchmark offers ample challenges for transfer learning approaches. We find the highest levels of zero-shot policy transfer from default Breakout agents to variants, occasionally with higher performance than variant-experts. However, transferred policies are generally far from optimal, which naturally motivates investigations into fast adaptation through transfer learning.
Overall, we observe substantial variance in the performance of fixed expert policies trained on defaults, then directly evaluated on all other variations.
Although game variants within curricula are related, the statistically significant interaction effects between factors suggest that they are indeed distinct tasks with substantial and diverse ``transfer gaps'', necessitating additional learning on top of default game policies. The most significant factors impacting zero-shot transfer were the ``difficulty'' switch in Space Invaders, the point accumulation ``Rules'' factor in Breakout, and ``Traffic'' thickness in Freeway, which we interpret as introducing the broadest differences. It is important to note that such modifications to default games do not easily fit within the more particular sets of assumptions often studied in transfer learning research. It is hard to argue that environment dynamics are precisely the same, or that state spaces are identical in distinct variants. Furthermore, even when observations change in subtle ways, e.g. the ``difficulty'' switch increases the size of the player's cannon in Space Invaders, this change has a measurable impact on agent transfer performance, and it is not immediately clear how domain adaptation approaches would mitigate the issue.
Hence, we go on to investigate how these diverse ``transfer gaps'' influence the performance of value-function finetuning, a more general transfer learning algorithm.

\begin{figure}[h]
\begin{center}
\includegraphics[width=0.99\textwidth]{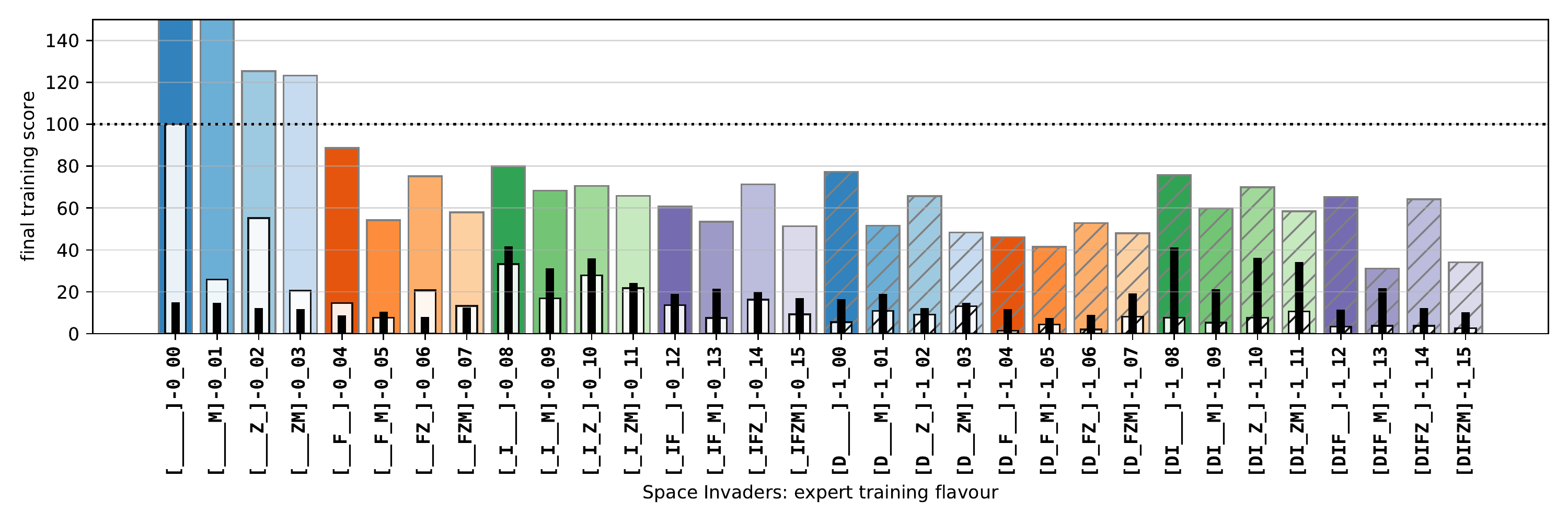}
\includegraphics[width=0.61\textwidth]{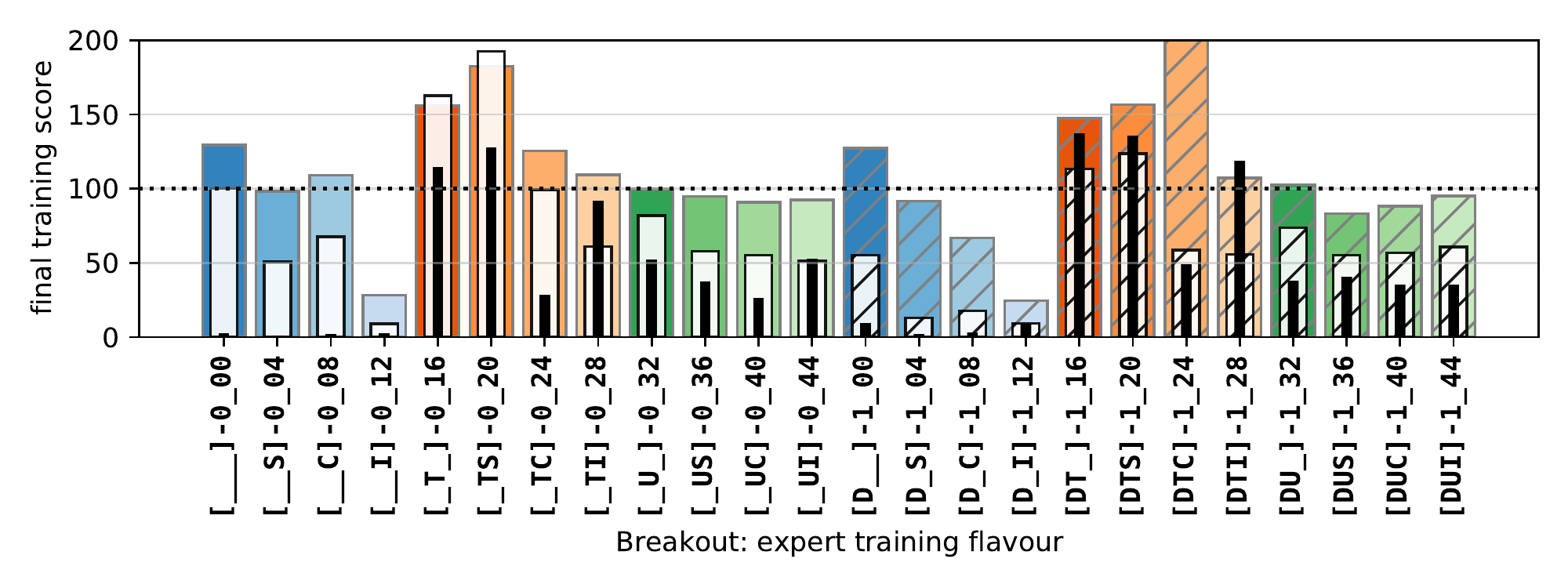}
\includegraphics[width=0.38\textwidth]{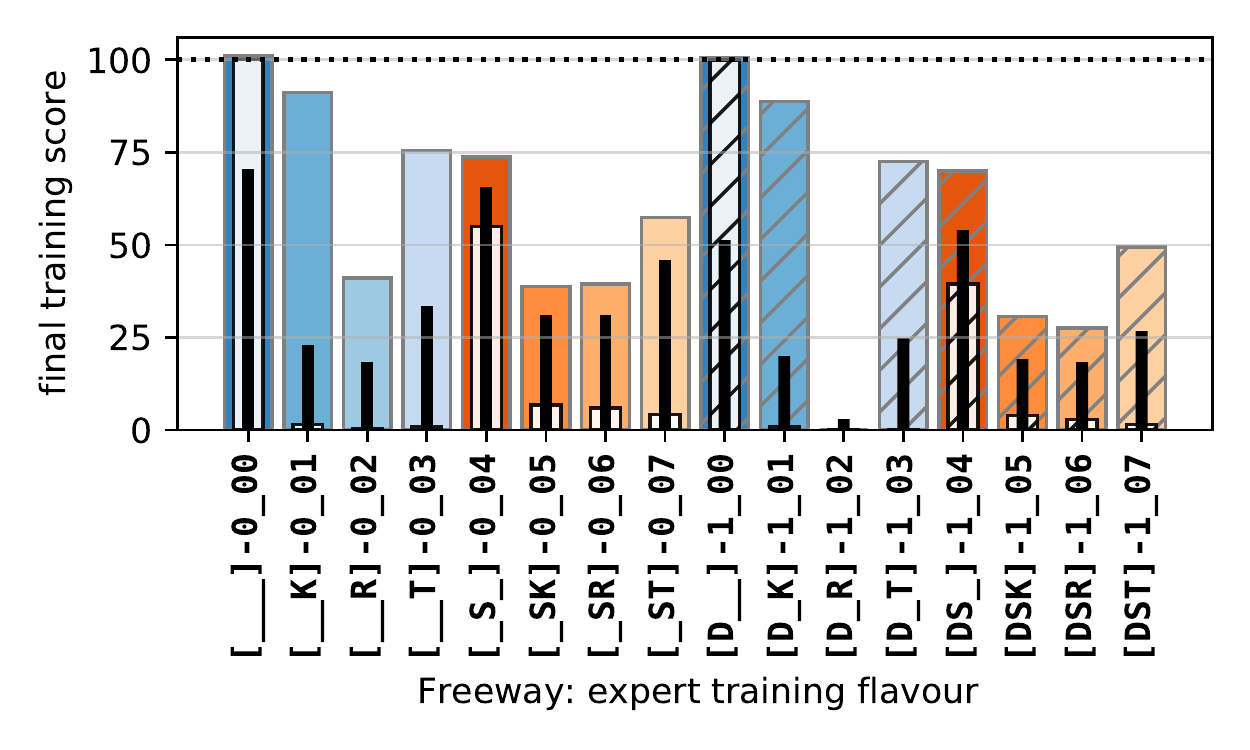}
\end{center}
\caption{Finetuning performance starting from default game expert $(0\_00)$ on all other variants, plotted as normalised scores (percentages) of the final performance of respective variant-experts, reported for Space Invaders (top), Breakout (bottom left) and Freeway variants (bottom right). Please note that $100\%$ indicates comparable performance with variant-experts, which were trained from scratch with $20\times$ more data than finetuning agents. Black bars show an ablation of finetuning training, starting from scratch rather than from default agent parameters. The delta to is due to transfer learning via agent network parameter initialisation. White bars indicate zero-shot transfer performance of default game agents. The delta to finetuning results represents the net benefit of learning in the transfer target variant. Hatching denotes the ``difficulty'' switch being on, and colours indicate the ``game mode'', see Figure~\ref{figures:variant_legend} for details.\label{figures:default_finetune}}
\end{figure}

\subsection{Finetuning of Default Game Experts}
Zero-shot policy transfer can be substantial, and as such, constitutes an appropriate starting point for fine-tuning on other variants. However, it is not immediately obvious that current RL agents can capitalise on this zero-shot policy transfer to meaningfully speed up learning relative to variant-experts.
In order to improve performance on new variants it important that agent representations and predictions are finetuned on data from transfer target tasks. We report results of default game expert finetuning experiments in Figure~\ref{figures:default_finetune}, together with the natural baseline of zero-shot policy transfer from the same experts, and an ablation: learning from scratch for the same amount of environment interaction as used for finetuning. In a majority of cases we find that default expert finetuning outperforms baseline and ablation approaches. Hence, a simple and general algorithm such as Rainbow-IQN finetuning can leverage knowledge acquired in default games to improve performance beyond the  zero-shot generalisation offered by its function approximator alone. Note that these gains are significant on the normalised scales induced by variant-expert performance, which account for the different learning dynamics caused by particular game variation, as shown in previous sections.

It is important not to overlook the heterogeneity of these finetuning results within and across curricula. Using a fixed, somewhat generous amount of data to finetune (10 million steps) does allow a substantial amount of performance to be acquired in some cases. Rainbow-IQN agents can also learn to a significant level of performance starting from scratch in this restricted training regime, which we would not have observed without the ablation experiment, e.g. on Freeway variants, but to some extent also on Breakout. Conversely, direct evaluation without training on the transfer target task can confer a competitive level of zero-shot performance, which we necessarily investigated in the previous section. Taken together, we interpret these heterogeneous finetuning results as strong motivation for research on improved general transfer learning algorithms, applicable across many transfer scenarios. 
But what can help transfer in this very general setting? Intuitively, learning several variants within curricula could offer a solution in the form of transfer using multiple sources, perhaps at the cost of additional complexity.

\begin{figure}[h]
\begin{center}
\includegraphics[width=0.42\textwidth]{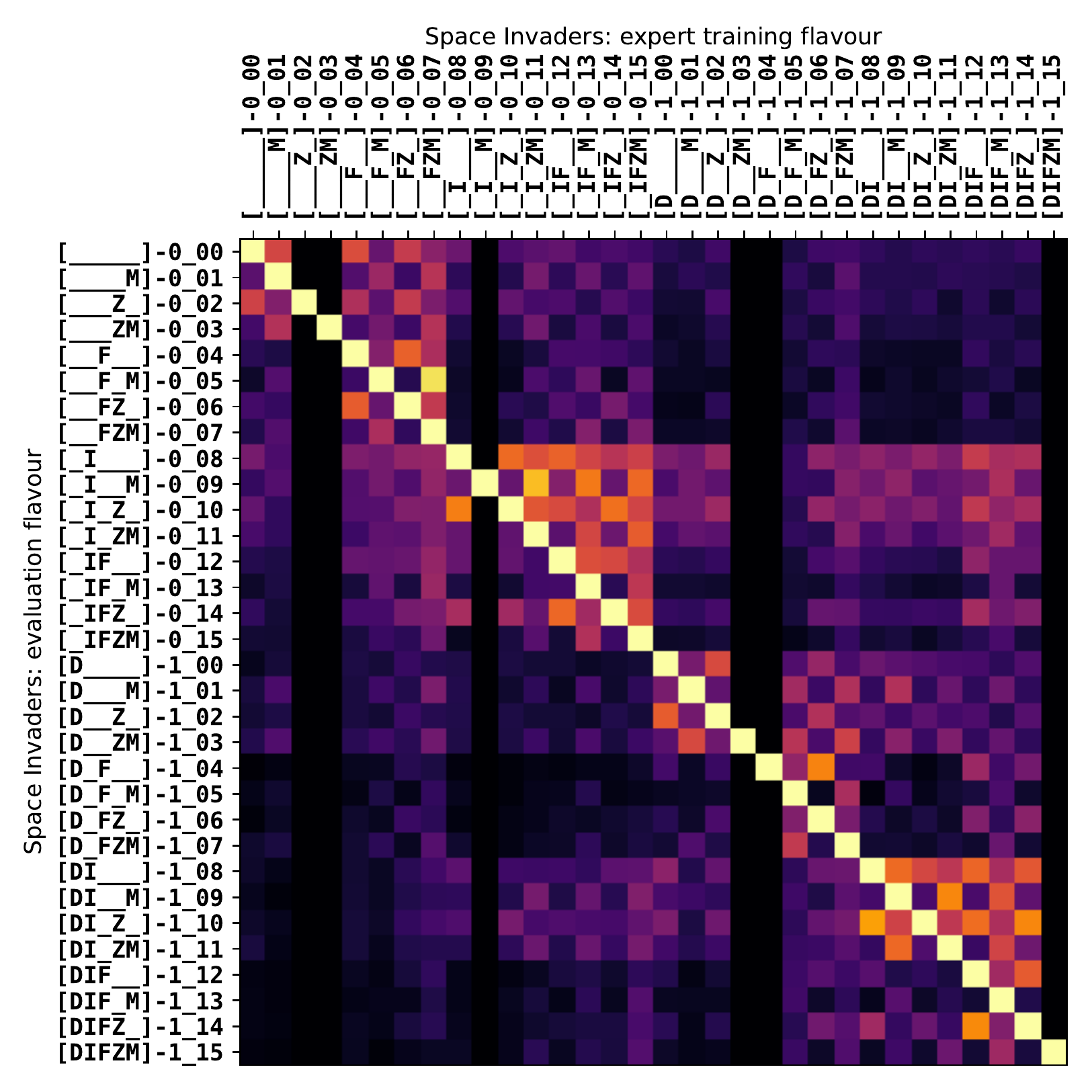}
\includegraphics[width=0.32\textwidth]{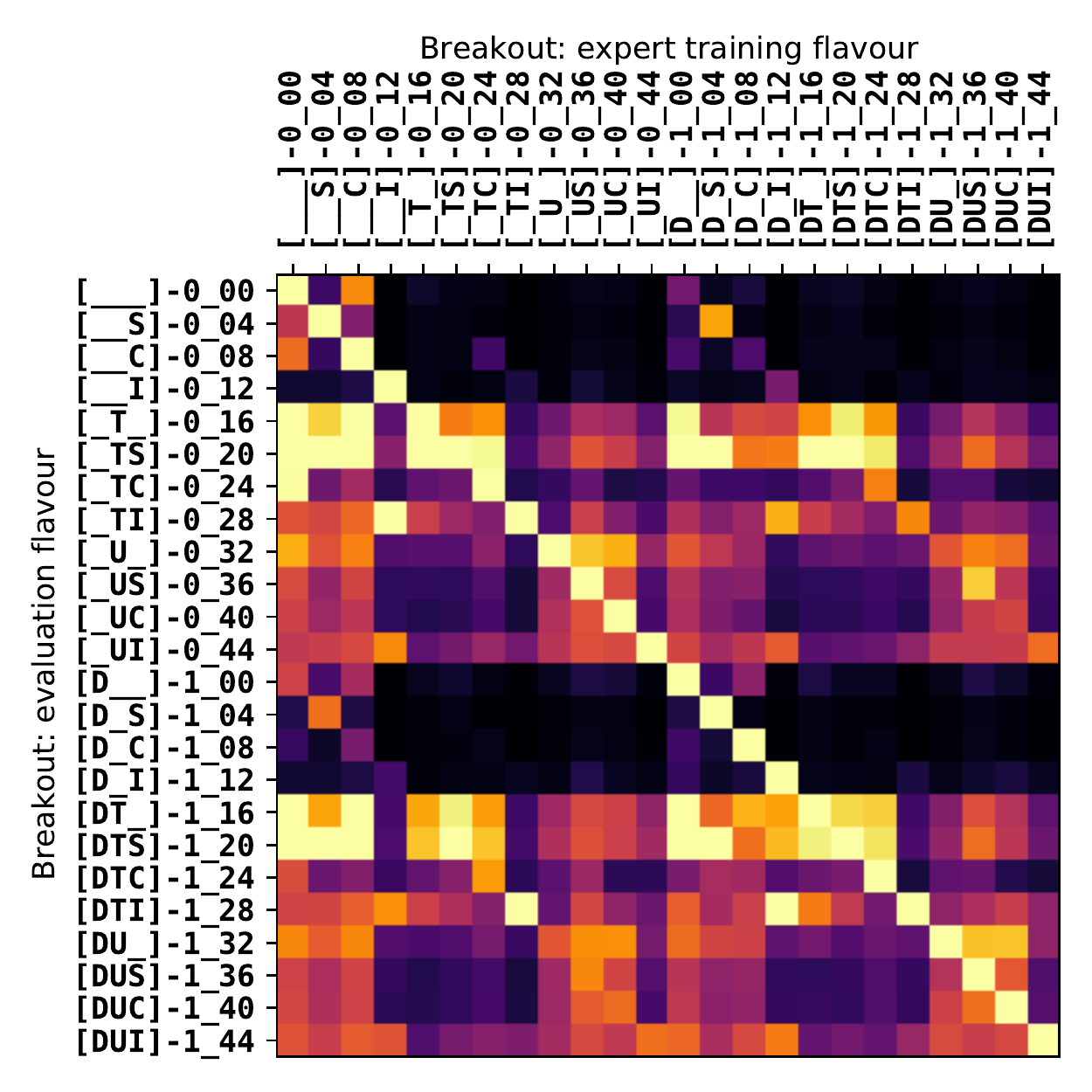}
\includegraphics[width=0.25\textwidth, trim=0 29 0 0]{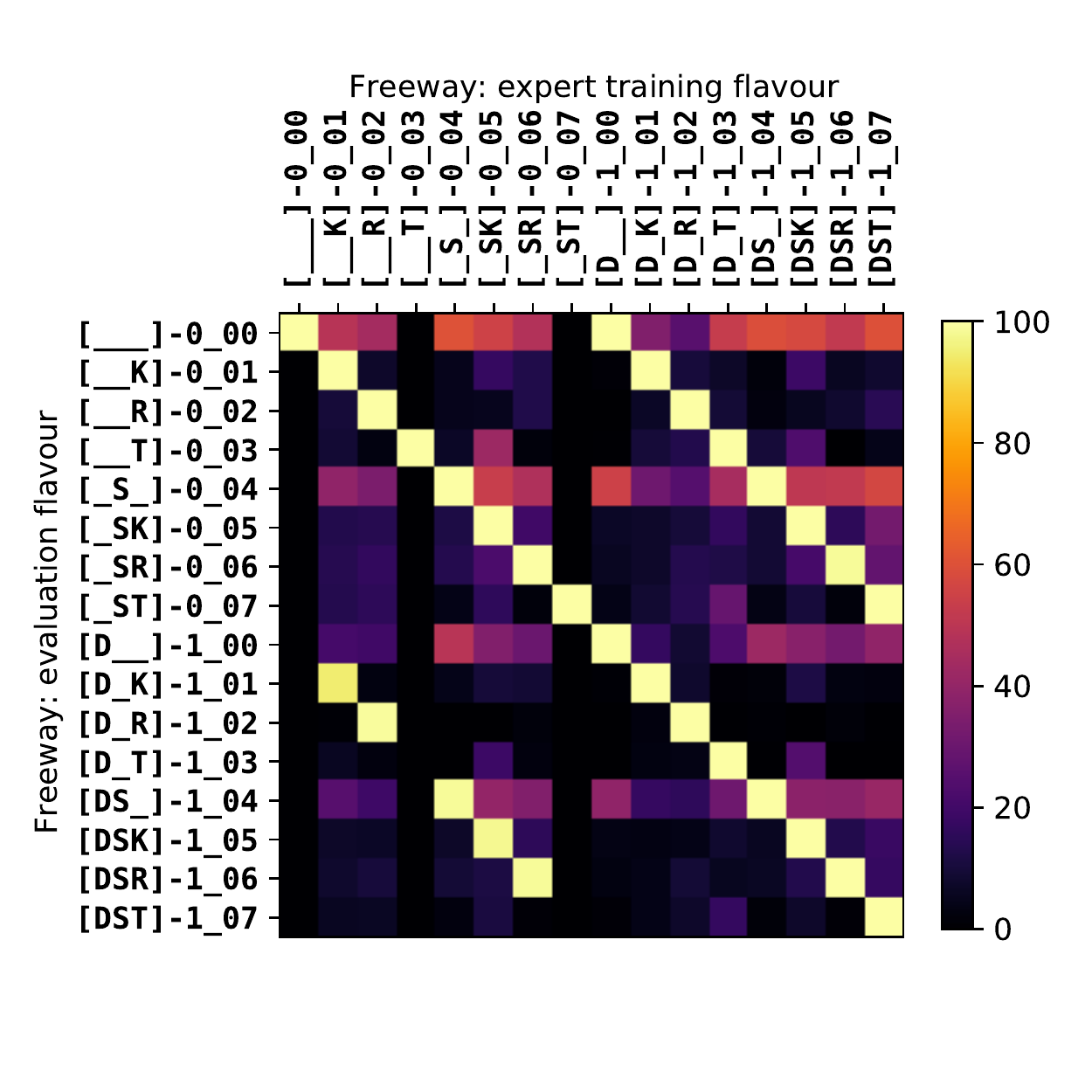}
\end{center}
\caption{Transfer matrices depicting zero-shot evaluation of Space Invaders (left), Breakout (middle) and Freeway variant-experts (right), plotted as columns, and evaluated on all variants (rows). Values reported as normalised scores (percentages) of respective variant-expert scores.\label{figure:jumpstart_tm}}
\end{figure}

\begin{figure}[h]
\begin{center}
\includegraphics[width=0.30\textwidth]{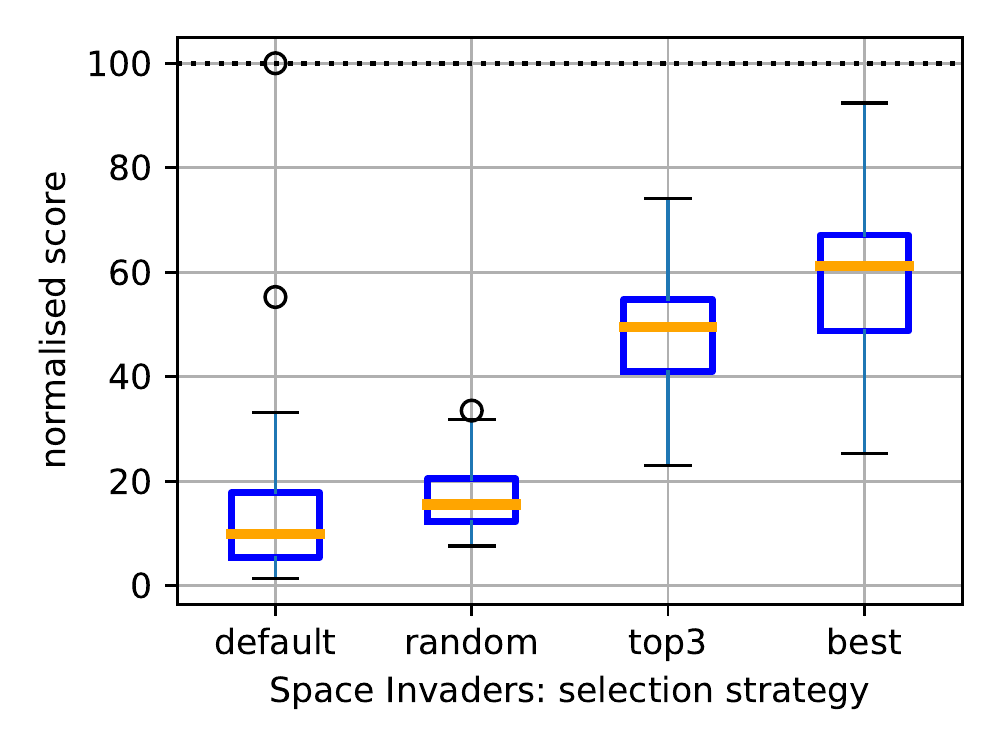}
\includegraphics[width=0.30\textwidth]{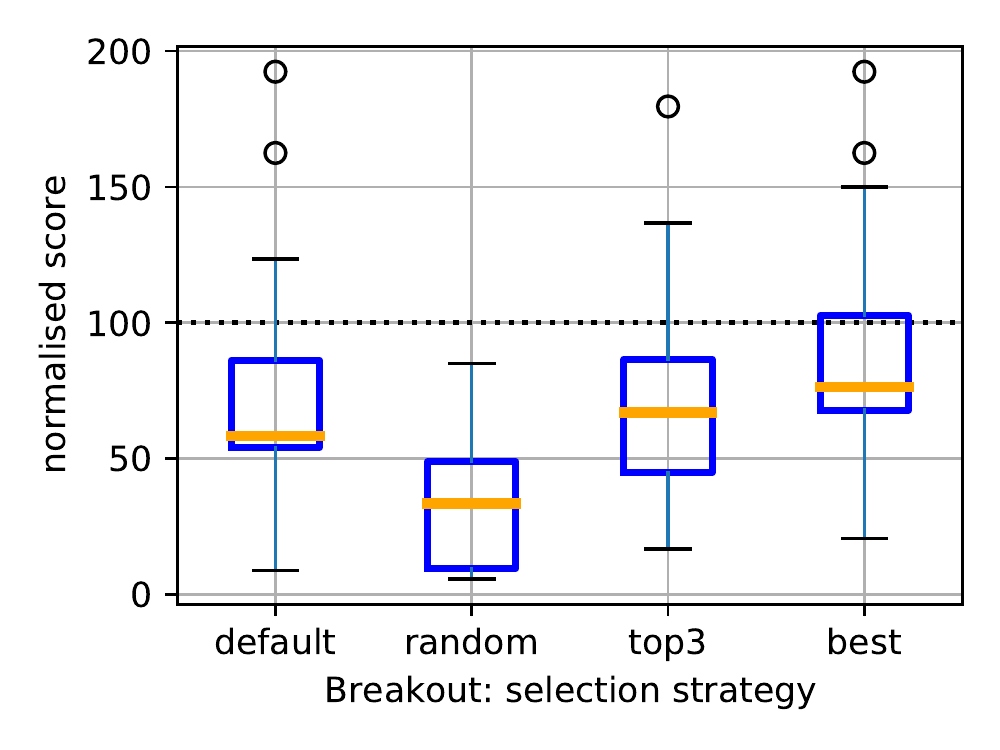}
\includegraphics[width=0.30\textwidth]{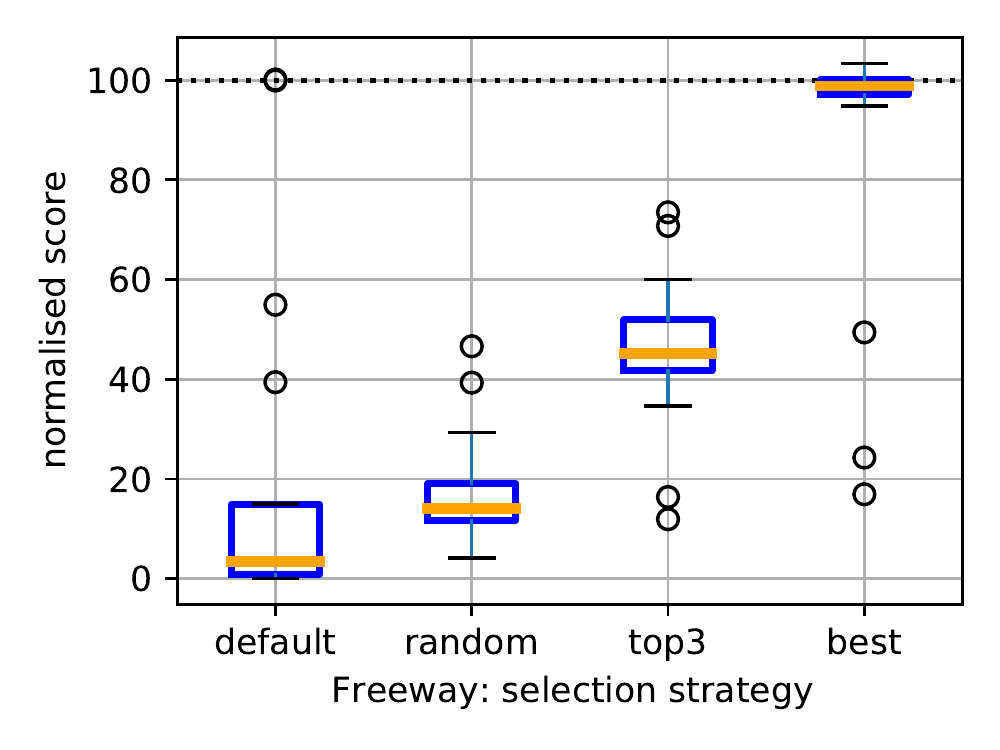}
\end{center}
\caption{Box-plots of zero-shot normalised score distributions for source task selection strategies on Space Invaders (left), Breakout (middle) and Freeway variants (right): ``default'' indicates using the default game expert ($0\_00$) policy to act in all variants; the following selection strategies denote ways of choosing policies exclusively from \emph{all other variants} in their respective domains: ``random'' means using a randomly chosen expert policy in every evaluation episode; ``top3'' and ``best'' strategies require access to some approximation of the transfer matrix to select the top three or best single source policy for each variant, from all other experts.\label{figure:jumpstart_box}}
\end{figure}

\subsection{Zero-shot Transfer Source Game Selection}

In this section, we focus on knowledge in the form of policies learned by experts on other game variants, and we perform an all-to-all zero-shot transfer evaluation, reported as transfer matrices in Figure~\ref{figure:jumpstart_tm}. We observe substantial off-diagonal structure in all games. The default game is not the best single source task in many cases, especially for variants with many factors introducing changes to the basic game. Naturally, we would like to know how well a superior source task selection strategy could perform on average. More precisely, we would like to empirically characterise the average upper bound on zero-shot transfer possible by appropriate selection of a single source policy for any game variant from all the remaining variant-expert policies.
We plot the expectation for success of several transfer source task selection strategies in Figure~\ref{figure:jumpstart_box}, as well as the ``best'' case selection strategy, which provides an upper bound for what can be achieved by selecting a single variant-expert's policy. For Freeway we observe strong transfer between policies learned with and without the ``difficulty'' switch being turned on. Matching the ``best'' strategy may be costly, e.g. in terms of expert evaluations, so we also report a ``top3'' selection strategy: zero-shot transfer performance of acting in every episode with one of the ``top3'' variant-experts at random. This strategy results in over $40\%$ median normalised performance for all curricula. This level of transfer could be achieved at the expense of some target task interactions for ascertaining the quality of all other variant-expert policies on the current game variation. Interestingly, for all game titles, the ``top3'' strategy is superior to zero-shot transfer from the default games. This cannot be said for the average or ``random'' choice strategy, which is inferior to zero-shot transfer from default experts in Breakout. These observations motivate further research into automatic selection of variant-expert policies for transfer. The rich off-diagonal structure also recommends the use of Atari game variations as systematic curricula in a continual learning setting, if only for leveraging the potential for policy reuse highlighted by our evaluations.

\section{Conclusions} 
Using Atari curricula, we have demonstrated that general transfer learning approaches, such as zero-shot policy generalisation and single expert model finetuning, lead to significant performance improvements in a large number of diverse transfer learning scenarios that are non-trivial for human players. Our analyses indicate that introducing modifications to basic games in a combinatorial fashion can substantially impact the empirical learning efficiency of a deep RL agent. Furthermore, zero-shot transfer performance from default games alone is also hindered by interactions between modifications.
These results highlight the value of learning several variant-experts and reusing their knowledge to quickly find effective policies for new variations. Such strategies have the potential to efficiently match expert performance on transfer target tasks, but require the ability to both accumulate knowledge across game variants, and to appropriately transfer from several source tasks at once. Our study serves as a proof-of-concept that Atari curricula offer an invaluable benchmark of suitable complexity for  systematically probing transfer without task engineering.

\bibliography{main}

\begin{thebibliography}{64}
\providecommand{\natexlab}[1]{#1}
\providecommand{\url}[1]{\texttt{#1}}
\expandafter\ifx\csname urlstyle\endcsname\relax
  \providecommand{\doi}[1]{doi: #1}\else
  \providecommand{\doi}{doi: \begingroup \urlstyle{rm}\Url}\fi

\bibitem[Barreto et~al.(2017)Barreto, Dabney, Munos, Hunt, Schaul, van Hasselt,
  and Silver]{Barreto2017successor}
Andre Barreto, Will Dabney, Remi Munos, Jonathan~J Hunt, Tom Schaul, Hado~P van
  Hasselt, and David Silver.
\newblock Successor features for transfer in reinforcement learning.
\newblock In I.~Guyon, U.~V. Luxburg, S.~Bengio, H.~Wallach, R.~Fergus,
  S.~Vishwanathan, and R.~Garnett (eds.), \emph{Advances in Neural Information
  Processing Systems}, volume~30. Curran Associates, Inc., 2017.
\newblock URL
  \url{https://proceedings.neurips.cc/paper/2017/file/350db081a661525235354dd3e19b8c05-Paper.pdf}.

\bibitem[Barreto et~al.(2019)Barreto, Borsa, Hou, Comanici, Ayg{\"u}n, Hamel,
  Toyama, Mourad, Silver, Precup, et~al.]{Barreto2019option}
Andr{\'e} Barreto, Diana Borsa, Shaobo Hou, Gheorghe Comanici, Eser Ayg{\"u}n,
  Philippe Hamel, Daniel Toyama, Shibl Mourad, David Silver, Doina Precup,
  et~al.
\newblock The option keyboard: Combining skills in reinforcement learning.
\newblock \emph{Advances in Neural Information Processing Systems}, 32, 2019.

\bibitem[{Bellemare} et~al.(2013){Bellemare}, {Naddaf}, {Veness}, and
  {Bowling}]{Bellemare13arcade}
M.~G. {Bellemare}, Y.~{Naddaf}, J.~{Veness}, and M.~{Bowling}.
\newblock The arcade learning environment: An evaluation platform for general
  agents.
\newblock \emph{Journal of Artificial Intelligence Research}, 47:\penalty0
  253--279, jun 2013.

\bibitem[Bellemare et~al.(2017)Bellemare, Dabney, and
  Munos]{Bellemare2017distributional}
Marc~G Bellemare, Will Dabney, and R{\'e}mi Munos.
\newblock A distributional perspective on reinforcement learning.
\newblock In \emph{International Conference on Machine Learning}, pp.\
  449--458. PMLR, 2017.

\bibitem[Bellman(1957)]{Bellman1957markovian}
Richard Bellman.
\newblock A markovian decision process.
\newblock \emph{Journal of mathematics and mechanics}, pp.\  679--684, 1957.

\bibitem[Bengio(2012)]{Bengio12transfer}
Yoshua Bengio.
\newblock Deep learning of representations for unsupervised and transfer
  learning.
\newblock In Isabelle Guyon, Gideon Dror, Vincent Lemaire, Graham Taylor, and
  Daniel Silver (eds.), \emph{Proceedings of ICML Workshop on Unsupervised and
  Transfer Learning}, volume~27 of \emph{Proceedings of Machine Learning
  Research}, pp.\  17--36, Bellevue, Washington, USA, 02 Jul 2012. PMLR.
\newblock URL \url{https://proceedings.mlr.press/v27/bengio12a.html}.

\bibitem[Berner et~al.(2019)Berner, Brockman, Chan, Cheung, Debiak, Dennison,
  Farhi, Fischer, Hashme, Hesse, et~al.]{Berner2019dota}
Christopher Berner, Greg Brockman, Brooke Chan, Vicki Cheung, Przemys{\l}aw
  Debiak, Christy Dennison, David Farhi, Quirin Fischer, Shariq Hashme, Chris
  Hesse, et~al.
\newblock Dota 2 with large scale deep reinforcement learning.
\newblock \emph{arXiv preprint arXiv:1912.06680}, 2019.

\bibitem[Caruana(1997)]{Caruana1997multitask}
Rich Caruana.
\newblock Multitask learning.
\newblock \emph{Machine learning}, 28\penalty0 (1):\penalty0 41--75, 1997.

\bibitem[Castro et~al.(2018)Castro, Moitra, Gelada, Kumar, and
  Bellemare]{Castro2018dopamine}
Pablo~Samuel Castro, Subhodeep Moitra, Carles Gelada, Saurabh Kumar, and Marc~G
  Bellemare.
\newblock Dopamine: A research framework for deep reinforcement learning.
\newblock \emph{arXiv preprint arXiv:1812.06110}, 2018.

\bibitem[Cobbe et~al.(2019)Cobbe, Klimov, Hesse, Kim, and
  Schulman]{Cobbe2019quantifying}
Karl Cobbe, Oleg Klimov, Chris Hesse, Taehoon Kim, and John Schulman.
\newblock Quantifying generalization in reinforcement learning.
\newblock In \emph{International Conference on Machine Learning}, pp.\
  1282--1289. PMLR, 2019.

\bibitem[Cobbe et~al.(2020)Cobbe, Hesse, Hilton, and
  Schulman]{Cobbe2020leveraging}
Karl Cobbe, Chris Hesse, Jacob Hilton, and John Schulman.
\newblock Leveraging procedural generation to benchmark reinforcement learning.
\newblock In \emph{International conference on machine learning}, pp.\
  2048--2056. PMLR, 2020.

\bibitem[Dabney et~al.(2018)Dabney, Ostrovski, Silver, and
  Munos]{Dabney2018iqn}
Will Dabney, Georg Ostrovski, David Silver, and R{\'e}mi Munos.
\newblock Implicit quantile networks for distributional reinforcement learning.
\newblock In \emph{International conference on machine learning}, pp.\
  1096--1105. PMLR, 2018.

\bibitem[Davidson et~al.(1993)Davidson, MacKinnon,
  et~al.]{davidson1993estimation}
Russell Davidson, James~G MacKinnon, et~al.
\newblock \emph{Estimation and inference in econometrics}, volume~63.
\newblock Oxford New York, 1993.

\bibitem[Farebrother et~al.(2018)Farebrother, Machado, and
  Bowling]{Farebrother2018generalization}
Jesse Farebrother, Marlos~C Machado, and Michael Bowling.
\newblock Generalization and regularization in dqn.
\newblock \emph{arXiv preprint arXiv:1810.00123}, 2018.

\bibitem[Fortunato et~al.(2018)Fortunato, Azar, Piot, Menick, Hessel, Osband,
  Graves, Mnih, Munos, Hassabis, et~al.]{Fortunato2018noisy}
Meire Fortunato, Mohammad~Gheshlaghi Azar, Bilal Piot, Jacob Menick, Matteo
  Hessel, Ian Osband, Alex Graves, Volodymyr Mnih, Remi Munos, Demis Hassabis,
  et~al.
\newblock Noisy networks for exploration.
\newblock In \emph{International Conference on Learning Representations}, 2018.

\bibitem[Girden(1992)]{girden1992anova}
Ellen~R Girden.
\newblock \emph{ANOVA: Repeated measures}.
\newblock Number~84. Sage, 1992.

\bibitem[Glass \& Hopkins(1996)Glass and Hopkins]{glass1996statistical}
Gene~V Glass and Kenneth~D Hopkins.
\newblock Statistical methods in education and psychology.
\newblock 1996.

\bibitem[Haarnoja et~al.(2018)Haarnoja, Zhou, Abbeel, and
  Levine]{Haarnoja2018soft}
Tuomas Haarnoja, Aurick Zhou, Pieter Abbeel, and Sergey Levine.
\newblock Soft actor-critic: Off-policy maximum entropy deep reinforcement
  learning with a stochastic actor.
\newblock In \emph{International conference on machine learning}, pp.\
  1861--1870. PMLR, 2018.

\bibitem[Hafner et~al.(2020)Hafner, Lillicrap, Norouzi, and
  Ba]{Hafner2020mastering}
Danijar Hafner, Timothy~P Lillicrap, Mohammad Norouzi, and Jimmy Ba.
\newblock Mastering atari with discrete world models.
\newblock In \emph{International Conference on Learning Representations}, 2020.

\bibitem[Hessel et~al.(2018)Hessel, Modayil, Van~Hasselt, Schaul, Ostrovski,
  Dabney, Horgan, Piot, Azar, and Silver]{Hessel2018rainbow}
Matteo Hessel, Joseph Modayil, Hado Van~Hasselt, Tom Schaul, Georg Ostrovski,
  Will Dabney, Dan Horgan, Bilal Piot, Mohammad Azar, and David Silver.
\newblock Rainbow: Combining improvements in deep reinforcement learning.
\newblock In \emph{Thirty-second AAAI conference on artificial intelligence},
  2018.

\bibitem[Hinton et~al.(2006)Hinton, Osindero, and Teh]{Hinton06}
Geoffrey~E. Hinton, Simon Osindero, and Yee~Whye Teh.
\newblock A fast learning algorithm for deep belief nets.
\newblock \emph{Neural Computation}, 18:\penalty0 1527--1554, 2006.

\bibitem[Horgan et~al.(2018)Horgan, Quan, Budden, Barth-Maron, Hessel, van
  Hasselt, and Silver]{Horgan2018distributed}
Dan Horgan, John Quan, David Budden, Gabriel Barth-Maron, Matteo Hessel, Hado
  van Hasselt, and David Silver.
\newblock Distributed prioritized experience replay.
\newblock In \emph{International Conference on Learning Representations}, 2018.

\bibitem[Hospedales et~al.(2020)Hospedales, Antoniou, Micaelli, and
  Storkey]{Hospedales2020meta}
Timothy~M. Hospedales, Antreas Antoniou, Paul Micaelli, and Amos~J. Storkey.
\newblock Meta-learning in neural networks: {A} survey.
\newblock \emph{CoRR}, abs/2004.05439, 2020.
\newblock URL \url{https://arxiv.org/abs/2004.05439}.

\bibitem[Jaderberg et~al.(2016)Jaderberg, Mnih, Czarnecki, Schaul, Leibo,
  Silver, and Kavukcuoglu]{Jaderberg2016reinforcement}
Max Jaderberg, Volodymyr Mnih, Wojciech~Marian Czarnecki, Tom Schaul, Joel~Z
  Leibo, David Silver, and Koray Kavukcuoglu.
\newblock Reinforcement learning with unsupervised auxiliary tasks.
\newblock \emph{arXiv preprint arXiv:1611.05397}, 2016.

\bibitem[Kingma \& Ba(2015)Kingma and Ba]{Kingma2015adam}
Diederik~P Kingma and Jimmy Ba.
\newblock Adam: A method for stochastic optimization.
\newblock In \emph{ICLR (Poster)}, 2015.

\bibitem[Levine \& Koltun(2013)Levine and Koltun]{Levine2013gps}
Sergey Levine and Vladlen Koltun.
\newblock Guided policy search.
\newblock In Sanjoy Dasgupta and David McAllester (eds.), \emph{Proceedings of
  the 30th International Conference on Machine Learning}, volume~28 of
  \emph{Proceedings of Machine Learning Research}, pp.\  1--9, Atlanta,
  Georgia, USA, 17--19 Jun 2013. PMLR.
\newblock URL \url{https://proceedings.mlr.press/v28/levine13.html}.

\bibitem[Levine et~al.(2016)Levine, Finn, Darrell, and Abbeel]{Levine2016end}
Sergey Levine, Chelsea Finn, Trevor Darrell, and Pieter Abbeel.
\newblock End-to-end training of deep visuomotor policies.
\newblock \emph{The Journal of Machine Learning Research}, 17\penalty0
  (1):\penalty0 1334--1373, 2016.

\bibitem[Lillicrap et~al.(2015)Lillicrap, Hunt, Pritzel, Heess, Erez, Tassa,
  Silver, and Wierstra]{Lillicrap2015continuous}
Timothy~P Lillicrap, Jonathan~J Hunt, Alexander Pritzel, Nicolas Heess, Tom
  Erez, Yuval Tassa, David Silver, and Daan Wierstra.
\newblock Continuous control with deep reinforcement learning.
\newblock \emph{arXiv preprint arXiv:1509.02971}, 2015.

\bibitem[Machado et~al.(2018)Machado, Bellemare, Talvitie, Veness, Hausknecht,
  and Bowling]{Machado2018revisiting}
Marlos~C Machado, Marc~G Bellemare, Erik Talvitie, Joel Veness, Matthew
  Hausknecht, and Michael Bowling.
\newblock Revisiting the arcade learning environment: Evaluation protocols and
  open problems for general agents.
\newblock \emph{Journal of Artificial Intelligence Research}, 61:\penalty0
  523--562, 2018.

\bibitem[Mnih et~al.(2013)Mnih, Kavukcuoglu, Silver, Graves, Antonoglou,
  Wierstra, and Riedmiller]{Mnih2013playing}
Volodymyr Mnih, Koray Kavukcuoglu, David Silver, Alex Graves, Ioannis
  Antonoglou, Daan Wierstra, and Martin Riedmiller.
\newblock Playing atari with deep reinforcement learning.
\newblock \emph{arXiv preprint arXiv:1312.5602}, 2013.

\bibitem[Mnih et~al.(2015)Mnih, Kavukcuoglu, Silver, Rusu, Veness, Bellemare,
  Graves, Riedmiller, Fidjeland, Ostrovski, et~al.]{Mnih2015dqn}
Volodymyr Mnih, Koray Kavukcuoglu, David Silver, Andrei~A Rusu, Joel Veness,
  Marc~G Bellemare, Alex Graves, Martin Riedmiller, Andreas~K Fidjeland, Georg
  Ostrovski, et~al.
\newblock Human-level control through deep reinforcement learning.
\newblock \emph{nature}, 518\penalty0 (7540):\penalty0 529--533, 2015.

\bibitem[Mnih et~al.(2016)Mnih, Badia, Mirza, Graves, Lillicrap, Harley,
  Silver, and Kavukcuoglu]{Mnih2016asynchronous}
Volodymyr Mnih, Adria~Puigdomenech Badia, Mehdi Mirza, Alex Graves, Timothy
  Lillicrap, Tim Harley, David Silver, and Koray Kavukcuoglu.
\newblock Asynchronous methods for deep reinforcement learning.
\newblock In \emph{International conference on machine learning}, pp.\
  1928--1937. PMLR, 2016.

\bibitem[OpenAI(2018)]{OpenAI2018compute}
OpenAI.
\newblock Ai and compute, 2018.
\newblock URL \url{https://openai.com/blog/ai-and-compute/}.

\bibitem[Pan \& Yang(2009)Pan and Yang]{Pan2009survey}
Sinno~Jialin Pan and Qiang Yang.
\newblock A survey on transfer learning.
\newblock \emph{IEEE Transactions on knowledge and data engineering},
  22\penalty0 (10):\penalty0 1345--1359, 2009.

\bibitem[Parisotto et~al.(2016)Parisotto, Ba, and
  Salakhutdinov]{Parisotto2016actor}
Emilio Parisotto, Lei~Jimmy Ba, and Ruslan Salakhutdinov.
\newblock Actor-mimic: Deep multitask and transfer reinforcement learning.
\newblock In \emph{ICLR (Poster)}, 2016.

\bibitem[Pascanu et~al.(2012)Pascanu, Mikolov, and
  Bengio]{Pascanu2012understanding}
Razvan Pascanu, Tom{\'{a}}s Mikolov, and Yoshua Bengio.
\newblock Understanding the exploding gradient problem.
\newblock \emph{CoRR}, abs/1211.5063, 2012.
\newblock URL \url{http://arxiv.org/abs/1211.5063}.

\bibitem[Puterman(1994)]{Puterman1994markov}
Martin~L Puterman.
\newblock Markov decision processes: Discrete stochastic dynamic programming,
  1994.

\bibitem[Riedmiller(2005)]{Riedmiller2005neural}
Martin Riedmiller.
\newblock Neural fitted q iteration--first experiences with a data efficient
  neural reinforcement learning method.
\newblock In \emph{European conference on machine learning}, pp.\  317--328.
  Springer, 2005.

\bibitem[Rusu et~al.(2016{\natexlab{a}})Rusu, Colmenarejo, G{\"u}l{\c{c}}ehre,
  Desjardins, Kirkpatrick, Pascanu, Mnih, Kavukcuoglu, and
  Hadsell]{Rusu2016distillation}
Andrei~A Rusu, Sergio~Gomez Colmenarejo, {\c{C}}aglar G{\"u}l{\c{c}}ehre,
  Guillaume Desjardins, James Kirkpatrick, Razvan Pascanu, Volodymyr Mnih,
  Koray Kavukcuoglu, and Raia Hadsell.
\newblock Policy distillation.
\newblock In \emph{ICLR (Poster)}, 2016{\natexlab{a}}.

\bibitem[Rusu et~al.(2016{\natexlab{b}})Rusu, Rabinowitz, Desjardins, Soyer,
  Kirkpatrick, Kavukcuoglu, Pascanu, and Hadsell]{Rusu2016progressive}
Andrei~A. Rusu, Neil~C. Rabinowitz, Guillaume Desjardins, Hubert Soyer, James
  Kirkpatrick, Koray Kavukcuoglu, Razvan Pascanu, and Raia Hadsell.
\newblock Progressive neural networks.
\newblock \emph{CoRR}, abs/1606.04671, 2016{\natexlab{b}}.
\newblock URL \url{http://arxiv.org/abs/1606.04671}.

\bibitem[Rusu et~al.(2017)Rusu, Večerík, Rothörl, Heess, Pascanu, and
  Hadsell]{Rusu2017sim2real}
Andrei~A. Rusu, Matej Večerík, Thomas Rothörl, Nicolas Heess, Razvan
  Pascanu, and Raia Hadsell.
\newblock Sim-to-real robot learning from pixels with progressive nets.
\newblock In Sergey Levine, Vincent Vanhoucke, and Ken Goldberg (eds.),
  \emph{Proceedings of the 1st Annual Conference on Robot Learning}, volume~78
  of \emph{Proceedings of Machine Learning Research}, pp.\  262--270. PMLR,
  13--15 Nov 2017.
\newblock URL \url{https://proceedings.mlr.press/v78/rusu17a.html}.

\bibitem[Schaul et~al.(2015)Schaul, Horgan, Gregor, and Silver]{Schaul15uva}
Tom Schaul, Daniel Horgan, Karol Gregor, and David Silver.
\newblock Universal value function approximators.
\newblock In Francis Bach and David Blei (eds.), \emph{Proceedings of the 32nd
  International Conference on Machine Learning}, volume~37 of \emph{Proceedings
  of Machine Learning Research}, pp.\  1312--1320, Lille, France, 07--09 Jul
  2015. PMLR.
\newblock URL \url{https://proceedings.mlr.press/v37/schaul15.html}.

\bibitem[Schaul et~al.(2016)Schaul, Quan, Antonoglou, and
  Silver]{Schaul2016prioritized}
Tom Schaul, John Quan, Ioannis Antonoglou, and David Silver.
\newblock Prioritized experience replay.
\newblock In \emph{ICLR (Poster)}, 2016.

\bibitem[Schaul et~al.(2018)Schaul, van Hasselt, Modayil, White, White, Bacon,
  Harb, Mourad, Bellemare, and Precup]{Schaul2018barbados}
Tom Schaul, Hado van Hasselt, Joseph Modayil, Martha White, Adam White,
  Pierre{-}Luc Bacon, Jean Harb, Shibl Mourad, Marc~G. Bellemare, and Doina
  Precup.
\newblock The barbados 2018 list of open issues in continual learning.
\newblock \emph{CoRR}, abs/1811.07004, 2018.
\newblock URL \url{http://arxiv.org/abs/1811.07004}.

\bibitem[Schmitt et~al.(2018)Schmitt, Hudson, Zidek, Osindero, Doersch,
  Czarnecki, Leibo, Kuttler, Zisserman, Simonyan,
  et~al.]{schmitt2018kickstarting}
Simon Schmitt, Jonathan~J Hudson, Augustin Zidek, Simon Osindero, Carl Doersch,
  Wojciech~M Czarnecki, Joel~Z Leibo, Heinrich Kuttler, Andrew Zisserman, Karen
  Simonyan, et~al.
\newblock Kickstarting deep reinforcement learning.
\newblock \emph{arXiv preprint arXiv:1803.03835}, 2018.

\bibitem[Schrittwieser et~al.(2020)Schrittwieser, Antonoglou, Hubert, Simonyan,
  Sifre, Schmitt, Guez, Lockhart, Hassabis, Graepel,
  et~al.]{Schrittwieser2020mastering}
Julian Schrittwieser, Ioannis Antonoglou, Thomas Hubert, Karen Simonyan,
  Laurent Sifre, Simon Schmitt, Arthur Guez, Edward Lockhart, Demis Hassabis,
  Thore Graepel, et~al.
\newblock Mastering atari, go, chess and shogi by planning with a learned
  model.
\newblock \emph{Nature}, 588\penalty0 (7839):\penalty0 604--609, 2020.

\bibitem[Schulman et~al.(2015)Schulman, Levine, Abbeel, Jordan, and
  Moritz]{Schulman2015trpo}
John Schulman, Sergey Levine, Pieter Abbeel, Michael Jordan, and Philipp
  Moritz.
\newblock Trust region policy optimization.
\newblock In Francis Bach and David Blei (eds.), \emph{Proceedings of the 32nd
  International Conference on Machine Learning}, volume~37 of \emph{Proceedings
  of Machine Learning Research}, pp.\  1889--1897, Lille, France, 07--09 Jul
  2015. PMLR.
\newblock URL \url{https://proceedings.mlr.press/v37/schulman15.html}.

\bibitem[Schulman et~al.(2017)Schulman, Wolski, Dhariwal, Radford, and
  Klimov]{Schulman2017proximal}
John Schulman, Filip Wolski, Prafulla Dhariwal, Alec Radford, and Oleg Klimov.
\newblock Proximal policy optimization algorithms.
\newblock \emph{arXiv preprint arXiv:1707.06347}, 2017.

\bibitem[Silver et~al.(2016)Silver, Huang, Maddison, Guez, Sifre, Van
  Den~Driessche, Schrittwieser, Antonoglou, Panneershelvam, Lanctot,
  et~al.]{Silver2016mastering}
David Silver, Aja Huang, Chris~J Maddison, Arthur Guez, Laurent Sifre, George
  Van Den~Driessche, Julian Schrittwieser, Ioannis Antonoglou, Veda
  Panneershelvam, Marc Lanctot, et~al.
\newblock Mastering the game of go with deep neural networks and tree search.
\newblock \emph{nature}, 529\penalty0 (7587):\penalty0 484--489, 2016.

\bibitem[Silver et~al.(2017)Silver, Schrittwieser, Simonyan, Antonoglou, Huang,
  Guez, Hubert, Baker, Lai, Bolton, et~al.]{Silver2017mastering}
David Silver, Julian Schrittwieser, Karen Simonyan, Ioannis Antonoglou, Aja
  Huang, Arthur Guez, Thomas Hubert, Lucas Baker, Matthew Lai, Adrian Bolton,
  et~al.
\newblock Mastering the game of go without human knowledge.
\newblock \emph{nature}, 550\penalty0 (7676):\penalty0 354--359, 2017.

\bibitem[Silver et~al.(2021)Silver, Singh, Precup, and
  Sutton]{Silver2021enough}
David Silver, Satinder Singh, Doina Precup, and Richard~S. Sutton.
\newblock Reward is enough.
\newblock \emph{Artificial Intelligence}, 299:\penalty0 103535, 2021.
\newblock ISSN 0004-3702.
\newblock \doi{https://doi.org/10.1016/j.artint.2021.103535}.
\newblock URL
  \url{https://www.sciencedirect.com/science/article/pii/S0004370221000862}.

\bibitem[Sutton \& Barto(2018)Sutton and Barto]{Sutton2018reinforcement}
Richard~S Sutton and Andrew~G Barto.
\newblock \emph{Reinforcement learning: An introduction}.
\newblock MIT press, 2018.

\bibitem[Taylor \& Stone(2009)Taylor and Stone]{Taylor2009transfer}
Matthew~E Taylor and Peter Stone.
\newblock Transfer learning for reinforcement learning domains: A survey.
\newblock \emph{Journal of Machine Learning Research}, 10\penalty0 (7), 2009.

\bibitem[Thrun \& Pratt(1998)Thrun and Pratt]{Thrun1998l2l}
Sebastian Thrun and Lorien Pratt.
\newblock \emph{Learning to Learn: Introduction and Overview}, pp.\  3--17.
\newblock Springer US, Boston, MA, 1998.
\newblock ISBN 978-1-4615-5529-2.
\newblock \doi{10.1007/978-1-4615-5529-2_1}.
\newblock URL \url{https://doi.org/10.1007/978-1-4615-5529-2_1}.

\bibitem[Tieleman \& Hinton(2012)Tieleman and Hinton]{Tieleman2012rmsprop}
Tijmen Tieleman and Geoffrey Hinton.
\newblock Rmsprop: Divide the gradient by a running average of its recent
  magnitude. coursera: Neural networks for machine learning.
\newblock \emph{COURSERA Neural Networks Mach. Learn}, 2012.

\bibitem[Tobin et~al.(2017)Tobin, Fong, Ray, Schneider, Zaremba, and
  Abbeel]{Tobin2017random}
Joshua Tobin, Rachel Fong, Alex Ray, Jonas Schneider, Wojciech Zaremba, and
  Pieter Abbeel.
\newblock Domain randomization for transferring deep neural networks from
  simulation to the real world.
\newblock \emph{CoRR}, abs/1703.06907, 2017.
\newblock URL \url{http://arxiv.org/abs/1703.06907}.

\bibitem[Toromanoff et~al.(2019)Toromanoff, Wirbel, and
  Moutarde]{Toromanoff2019rainbowiqn}
Marin Toromanoff, {\'{E}}milie Wirbel, and Fabien Moutarde.
\newblock Is deep reinforcement learning really superhuman on atari?
\newblock \emph{CoRR}, abs/1908.04683, 2019.
\newblock URL \url{http://arxiv.org/abs/1908.04683}.

\bibitem[Van~Hasselt et~al.(2016)Van~Hasselt, Guez, and Silver]{Van2016deep}
Hado Van~Hasselt, Arthur Guez, and David Silver.
\newblock Deep reinforcement learning with double q-learning.
\newblock In \emph{Proceedings of the AAAI conference on artificial
  intelligence}, volume~30, 2016.

\bibitem[Vinyals et~al.(2019)Vinyals, Babuschkin, Czarnecki, Mathieu, Dudzik,
  Chung, Choi, Powell, Ewalds, Georgiev, et~al.]{Vinyals2019grandmaster}
Oriol Vinyals, Igor Babuschkin, Wojciech~M Czarnecki, Micha{\"e}l Mathieu,
  Andrew Dudzik, Junyoung Chung, David~H Choi, Richard Powell, Timo Ewalds,
  Petko Georgiev, et~al.
\newblock Grandmaster level in starcraft ii using multi-agent reinforcement
  learning.
\newblock \emph{Nature}, 575\penalty0 (7782):\penalty0 350--354, 2019.

\bibitem[Wang et~al.(2016)Wang, Bapst, Heess, Mnih, Munos, Kavukcuoglu, and
  de~Freitas]{Wang2016sample}
Ziyu Wang, Victor Bapst, Nicolas Heess, Volodymyr Mnih, Remi Munos, Koray
  Kavukcuoglu, and Nando de~Freitas.
\newblock Sample efficient actor-critic with experience replay.
\newblock 2016.

\bibitem[Watkins \& Dayan(1992)Watkins and Dayan]{Watkins1992q}
Christopher~JCH Watkins and Peter Dayan.
\newblock Q-learning.
\newblock \emph{Machine learning}, 8\penalty0 (3):\penalty0 279--292, 1992.

\bibitem[Watkins(1989)]{Watkins1989learning}
Christopher John Cornish~Hellaby Watkins.
\newblock Learning from delayed rewards.
\newblock 1989.

\bibitem[Yu et~al.(2020)Yu, Quillen, He, Julian, Hausman, Finn, and
  Levine]{Yu2020metaworld}
Tianhe Yu, Deirdre Quillen, Zhanpeng He, Ryan Julian, Karol Hausman, Chelsea
  Finn, and Sergey Levine.
\newblock Meta-world: A benchmark and evaluation for multi-task and meta
  reinforcement learning.
\newblock In \emph{Conference on Robot Learning}, pp.\  1094--1100. PMLR, 2020.

\bibitem[Zhu et~al.(2020)Zhu, Lin, and Zhou]{Zhu2020survey}
Zhuangdi Zhu, Kaixiang Lin, and Jiayu Zhou.
\newblock Transfer learning in deep reinforcement learning: {A} survey.
\newblock \emph{CoRR}, abs/2009.07888, 2020.
\newblock URL \url{https://arxiv.org/abs/2009.07888}.

\end{thebibliography}
\bibliographystyle{collas2022_conference}

\newpage
\appendix

\section{Appendix: Extended Background}
\label{appendix:extended_background}

{\bf Q-learning.} \citet{Watkins1989learning} propose an RL algorithm which iteratively improves a parametric estimate $\hat Q_{\theta}~=~\hat Q_{\mdp}(\cdot, \cdot, \theta)$ of the optimal action-value function $Q_{\mdp}^{*}$ using the Bellman optimality operator \citep{Bellman1957markovian}: $$\hat Q_{\theta}(x, a)~\leftarrow~\expectation{\transitions}{\rewards(x, a, x')} + \gamma \expectation{\transitions}{\max_{a' \in \actions} \hat Q_{\theta}(x', a')}.$$

In model-free RL we do not assume explicit knowledge of the transition function $\transitions$, but the above optimisation can be implemented using samples procured by acting greedily with respect to the estimate $\hat Q_{\theta}$ with probability $1 - \epsilon$, and choosing actions uniformly at random otherwise. \citet{Watkins1992q} show that, under certain condition, this procedure converges to $Q^{*}_{\mdp}$, and an optimal policy can be derived as $\policy^{*}_{\mdp} (x) = \argmax{a \in \actions} \hat Q_{\theta}(x, a)$.

{\bf Deep Q-Learning.} \citet{Mnih2015dqn} use a convolutional neural network to parameterise $\hat Q_{\theta}$, and train the resulting Deep Q-Network (DQN) by iterative minimisation of the squared temporal difference (TD) error: 

Samples $(x_t, a_t, r_t, x_{t+1})$ are drawn randomly from a moving-window replay memory, where experience is collected online, and $\theta^{-}$ denotes a slow moving copy of online parameters $\theta$. DQN agents are efficiently trained using RMSProp \citep{Tieleman2012rmsprop}, a variant of mini-batch stochastic gradient descent (SGD).  This approach, with fixed hyper-parameters, was the first to master many classic Atari 2600 games from the Arcade Learning Environment (ALE) \citep{Bellemare13arcade} up to a casual human level of performance, using screen pixels as inputs. Several improvements are brought together into the Rainbow agent by \citet{Hessel2018rainbow}. \citet{Dabney2018iqn} introduce Implicit Quantile Networks (IQN), a principled generalisation of DQN such that the full return distribution $Z_{\mdp}^{\policy}$ is modelled implicitly through its quantile function, leading to superior performance. Rainbow-IQN was shown to be state-of-the-art at the time for the standard ($200$ million steps) data regime by \citet{Toromanoff2019rainbowiqn}. This family of algorithms remains the most widely implemented of deep RL approaches.

{\bf Deep Reinforcement Learning.} Online model-free RL with deep neural networks has been known to suffer from learning instability due to correlations between parameter updates. Stabilisation heuristics proved successful early on, such as off-line, full-batch training \citep{Riedmiller2005neural, Schulman2015trpo}, using a replay memory to decorrelate parameter updates \citep{Mnih2013playing, Schaul2016prioritized}, and correcting over-estimation errors \citep{Van2016deep}. \citet{Mnih2016asynchronous} introduced A3C, an online deep RL algorithm which leveraged parallel data generation for stable learning, further improved with auxiliary losses \citep{Jaderberg2016reinforcement}, and a replay memory \citep{Wang2016sample} for better sample efficiency. Deep RL has also been successfully applied to continuous control domains \citep{Lillicrap2015continuous, Schulman2015trpo, Schulman2017proximal, Haarnoja2018soft}. Model-based RL algorithms have proven successful in full information environments which require complex strategy, such as the game of Go \citep{Silver2016mastering, Silver2017mastering}, several board-games \citep{Schrittwieser2020mastering} and domains with limited data, such as robotic control \citep{Levine2013gps, Levine2016end}.  Recently, model-based RL approaches with learned models have proven competitive with model-free algorithms on domains with high-dimensional observations, e.g. DreamerV2 \citep{Hafner2020mastering}, and MuZero \citep{Schrittwieser2020mastering}. More complex training strategies have also been successful, such as combining deep RL with supervised learning from human games \citep{Silver2016mastering}, and using self-play, or multi-agent training, leading to human Grandmaster level performance in Starcraft II \citep{Vinyals2019grandmaster}, and winning against world champions at an e-sports game \citep{Berner2019dota}. It is important to note that such results are not yet achievable without access to super-computing infrastructure, e.g. thousands of GPUs or TPUs, and complex software stacks for distributed training, with estimated costs in single digit millions of dollars \citep{OpenAI2018compute}. Even when such resources are available, agent training can still take many months, so detailed ablations or hyper-parameter tuning become prohibitively expensive \citep{Berner2019dota}. Hence, improving data and computational efficiency through principled algorithmic innovations is essential for state-of-the-art deep RL.

\section{Appendix: Agent Training Details}
\label{appendix:agent_training_details}

We tuned relevant hyper-parameters for each game variant in order to avoid sub-optimal performance due to poor settings. Please note that we are the first to report Rainbow-IQN results on Atari variants, and hence we had only the DQN experiments of \citet{Farebrother2018generalization} as a guide, but no results were reported for the standard $200$ million steps training regime, and the two algorithms have many differences which can impact learning dynamics. \citet{Farebrother2018generalization} also used weight decay and dropout to reduce parameter norms and improve the generalisation of trained agents, but final performance was negatively impacted. We share the intuition that models which are closer to a random initialisation have an increased chance to generalise to related tasks, and may even be easier to adapt to new variants during finetuning.

\begin{table}[h]
\caption{Variant-Expert Rainbow-IQN hyper-parameters. When multiple values are specified, the final hyper-parameter was chosen via grid search. Note that each configuration was trained using two different random seeds.}
\centering
\label{tables:expert_training_hypers}
\vspace{10pt}
\begin{tabular}{L{4.5cm}C{4cm}L{6cm}}
{\bf Name}                                          &   {\bf Value(s)}          &   {\bf Units/Comments/Attributions}    \\ \hline
Reward Clipping Bound                               &   $1$                     &   Possible values $\{-1, 0, 1\}$ \citep{Mnih2015dqn}. \\
Replay Capacity                                     &   $\num{1000000}$         &   Transitions. \citep{Mnih2015dqn}.            \\
Replay Initial Size                                 &   $\num{500000}$          &   Transitions. \citep{Hessel2018rainbow}.\\
Initial Policy                                      &   $\epsilon = 1$          &   Replay memory initialised using random policy. \\
Initial Policy $\epsilon$ Annealing Steps           &   $\num{100000}$          &   Linearly annealed to $\epsilon = 0.01$ \citep{Mnih2015dqn}. \\
Behaviour Policy ($\epsilon$-greedy)                &   $\epsilon = 0.01$       &   The same during training and evaluation.  \\
Action Repeats                                      &   $4$                     &   Environment steps. \citep{Mnih2015dqn}.    \\
History Length                                      &   $4$                     &   Agent steps.    \citep{Mnih2015dqn}.    \\
Discount Factor                                     &   $\gamma = 0.99$         &   \citet{Mnih2015dqn}. \\
Multi-step Bootstrap                                &   $n = 3$                 &   \citet{Hessel2018rainbow}. \\
Double Q-Learning                                   &   True                    &   \citet{Van2016deep}.     \\ \hline
Prioritised Replay                                  &   True                    &   \citet{Schaul2016prioritized, Hessel2018rainbow}.  \\
Prioritisation Exponent                             &   $0.6$                   &   \citet{Horgan2018distributed}.  \\
Prioritisation Type                                 &   proportional            &   \citet{Hessel2018rainbow}.  \\
Prioritisation Importance Sampling                  &   $0.4$                   &   \citet{Horgan2018distributed}. \\ \hline
Parallel Environments                               &   $1$                     &   \citet{Toromanoff2019rainbowiqn}. \\    
Batch size                                          &   $\{32, 64\}$            &   \citet{Mnih2015dqn} \\
Optimiser                                           &   Clipped-SGD             &   Re-scaling gradients with norms larger than $C$. \\
Gradient Clipping Norm                              &   $C = 10$                &   Largest admissible gradient norm.   \\
Agent Steps per Update                              &   $\{4, 8\}$              &   Agent steps \citep{Mnih2015dqn}. \\ 
Learning Rate                                       &   $\{\num{d-6}, \num{d-5}, \num{d-4}, \num{d-3}\}$    &  \citet{Hessel2018rainbow}.    \\
Target Network Update Period                        &  $\{\num{1000}, \num{10000}\}$    &  Agent steps \citep{Castro2018dopamine}.  \\ \hline
Number of $\tau$ Samples                            &   $N = 64$                          &   \citet{Dabney2018iqn}.  \\
Number of $\tau'$ Samples                           &   $N'= 64$                         &   \citet{Dabney2018iqn}.  \\
Number of $\tilde \tau$ Samples (Policy)            &   $K = 64$                        &   \citet{Dabney2018iqn}.  \\
\end{tabular}
\end{table}

We replaced RMSProp \citep{Hinton06}, used by DQN \citep{Mnih2015dqn}, and Adam \citep{Kingma2015adam}, used by Rainbow \citep{Hessel2018rainbow}, with Stochastic Gradient Descent (SGD) and gradient norm clipping \citep{Pascanu2012understanding}. Please not that the aforementioned optimisers can provide superior performance, but tend to optimistically increase learning rates when parameter updates are correlated. We used hyper-parameter grid search to choose appropriate learning rates, and we purposefully also including smaller than typical values in the grid; we also allowed larger batch sizes and less frequent parameter updates besides standard values, see Table~\ref{tables:expert_training_hypers} for details.  To further avoid parameter norms increasing due to correlated updates at the start of training, we sampled $\num{500000}$ transitions in the replay memory before starting learning, using $\epsilon$ decay from fully random behaviour to an $\epsilon$-greedy policy with $\epsilon = 0.01$. The intuition behind these choices is that sampling data using a diverse set of behaviours policies will reduce the chances of correlated updates and learning divergence early on in training.

\begin{table}[h]
\caption{Rainbow-IQN expert finetuning hyper-parameters. When multiple values are specified, the final hyper-parameter was chosen via grid search. Note that each configuration was trained using two different random seeds.}
\centering
\label{tables:expert_finetuning_hypers}
\vspace{10pt}
\begin{tabular}{L{4.5cm}C{4cm}L{6cm}}
{\bf Name}                                          &   {\bf Value(s)}          &   {\bf Units/Comments/Attributions}    \\ \hline
Reward Clipping Bound                               &   $1$                     &   Possible values $\{-1, 0, 1\}$ \citep{Mnih2015dqn}. \\
Replay Capacity                                     &   $\num{1000000}$         &   Transitions. \citep{Mnih2015dqn}.            \\
Replay Initial Size                                 &   $\num{500000}$          &   Transitions. \citep{Hessel2018rainbow}.\\
Initial Policy                                      &   $\epsilon = 1$          &   Replay memory initialised using random policy. \\
Initial Policy $\epsilon$ Annealing                 &   $\num{100000}$          &   Linearly annealed to $\epsilon = 0.01$ \citep{Mnih2015dqn}. \\
Behaviour Policy ($\epsilon$-greedy)                &   $\epsilon = 0.01$       &   The same during training and evaluation.  \\
Action Repeats                                      &   $4$                     &   Environment steps. \citep{Mnih2015dqn}.    \\
History Length                                      &   $4$                     &   Agent steps.    \citep{Mnih2015dqn}.    \\
Discount Factor                                     &   $\gamma = 0.99$         &   \citet{Mnih2015dqn}. \\
Multi-step Bootstrap                                &   $n = 3$                 &   \citet{Hessel2018rainbow}. \\
Double Q-Learning                                   &   True                    &   \citet{Van2016deep}.     \\ \hline
Prioritised Replay                                  &   True                    &   \citet{Schaul2016prioritized, Hessel2018rainbow}.  \\
Prioritisation Exponent                             &   $0.6$                   &   \cite{Horgan2018distributed}.  \\
Prioritisation Type                                 &   proportional            &   \citet{Hessel2018rainbow}.  \\
Prioritisation Importance Sampling                  &   $0.4$                   &   \citet{Horgan2018distributed}. \\ \hline
Parallel Environments                               &   $1$                     &   \citet{Toromanoff2019rainbowiqn}. \\    
Batch size                                          &   $32$                    &   \citet{Mnih2015dqn} \\
Optimiser                                           &   Clipped-SGD             &   Re-scaling gradients with norms larger than $C$. \\
Gradient Clipping Norm                              &   $C \in \{0, 10, 40\}$   &   Largest admissible gradient norm. Setting $C=0$ disables gradient clipping. \\
Agent Steps per Update                              &   $4$                     &   Agent steps \citep{Mnih2015dqn}. \\
Learning Rate                                       &   $\{\num{d-7}, \num{d-6}, \num{5d-6}$, $\num{d-5}, \num{5d-5}, \num{d-4}\}$    &  \citet{Hessel2018rainbow}.    \\
Target Network Update Period                        &   $\num{10000}$           &  Agent steps \citep{Castro2018dopamine}.  \\ \hline
Number of $\tau$ Samples                            &   $N = 64$                  &   \citet{Dabney2018iqn}.  \\
Number of $\tau'$ Samples                           &   $N' = 64$                 &   \citet{Dabney2018iqn}.  \\
Number of $\tilde \tau$ Samples (Policy)            &   $K = 64$                &   \citet{Dabney2018iqn}.  \\
\end{tabular}
\end{table}

We used a slightly modified hyper-parameter grid for expert finetuning, primarily aimed at finding appropriate learning rates and the optimal levels of gradient clipping, as shown in Table~\ref{tables:expert_finetuning_hypers}. Expert finetuning is performed using a total of $10$ million environment steps, which is $20\times$ less experience and parameter updates compared to expert training from scratch. Hence, we allowed for more lenient gradient clipping, as well as disabling it altogether.

\begin{table}[h]
\caption{Rainbow-IQN Network Architecture.}
\centering
\label{tables:expert_architecture}
\vspace{10pt}
\begin{tabular}{ll}
{\bf Name}                  &   {\bf Value(s)}  \\ \hline
Convolution Channels        &   $16, 32, 32$   \\
Convolution Filter Sizes    &   $8\times8, 4\times4, 3\times3$   \\
Convolution Filter Stride   &   $4\times4, 2\times2, 1\times1$   \\
Convolution Activations     &   ReLU, ReLU, Identity   \\
Embedding Hiddens           &   $1568$   \\ 
Embedding Activations       &   ReLU   \\ 
Latent Dimension            &   $64$   \\
Combined Hiddens            &   $256, 18$   \\
Combined Activations        &   ReLU, Identity \\ 
\hline
\end{tabular}
\end{table}

Our choice to use hyper-parameter grid searches for both expert training and finetuning has greatly increased the computational costs of this study. We have somewhat offset these costs by reducing the numbers of filters in the convolutional trunk of the agent network, to the same sizes used by \citet{Mnih2013playing}, but we kept the number of convolutional layers at $3$ \citep{Mnih2015dqn, Dabney2018iqn}. Other network parameters were set according to \citet{Dabney2018iqn}, see architecture details in Table~\ref{tables:expert_architecture}. These modifications did not significantly impact agent performance, which is largely comparable to results reported by \citet{Castro2018dopamine} on default games.

\section{Appendix: Statistical Analyses}
\label{appendix:statistical_analyses}

We used generalised linear models, fit with robust covariance estimates (`HC3'), to perform multi-factor Analyses of Variance (ANOVA). After rejecting null hypotheses of the omnibus tests, we used Bonferroni corrections for multiple comparisons to perform post-hoc analyses of differences between group means. The desired significance level was $\alpha = 0.05$, which we corrected to $\alpha_c = 0.0015$ for Space Invaders, $\alpha_c = 0.0020$, for Breakout, $\alpha_c = 0.0031$ for Freeway, since the numbers of groups are $32$, $24$ and $16$ respectively.

\begin{table}[h]
\caption{Multi-Factor ANOVA (type 3) of Expert Scores vs. design factors, extended tables.}
\centering
\label{tables:anova_experts_full}
\vspace{10pt}
\begin{tabular}{lcccc}
\emph{Space Invaders}       &   $sum\_sq$       &   $df$        &   $F$             &   $PR(>F)$        \\ \hline
Intercept			    	&   $8.18e+07$      &   $1$         &   $10862.43$      &   $3.67e-73$      \\ \hline
{\bf Difficulty}		    &   $8.73e+05$      &   $1$         &   $  115.90$      &   $5.32e-16$      \\
{\bf Invisible Invaders}	&   $7.54e+06$      &   $1$         &   $ 1000.59$      &   $8.67e-41$      \\
{\bf Fast Bombs}		    &   $2.37e+06$      &   $1$         &   $  314.09$      &   $2.24e-26$      \\
{\bf Zigzagging Bombs}		&   $9.10e+04$      &   $1$         &   $   12.08$      &   $9.22e-04$      \\
{\bf Moving Shields}	    &   $9.46e+05$      &   $1$         &   $  125.51$      &   $9.93e-17$      \\ \hline
{\bf Interaction} 		    &   $2.63e+08$      &   $26$        &   $ 1341.37$      &   $5.33e-78$      \\ \hline
Residual                    &   $4.82e+05$      &   $64$        &                   &                   \\
\\
\emph{Breakout}             &   $sum\_sq$       &   $df$        &   $F$             &   $PR(>F)$        \\ \hline
Intercept				    &   $4.06e+06$      &   $1$         &   $7623.13$       &   $1.49e-54$      \\ \hline
{\bf Difficulty}		    &   $2.99e+04$      &   $1$         &   $  56.28$       &   $1.26e-09$      \\
{\bf Rules}			        &   $1.46e+06$      &   $2$         &   $1368.66$       &   $4.71e-43$      \\
{\bf Extras}			    &   $2.87e+05$      &   $3$         &   $ 179.84$       &   $4.22e-26$      \\ \hline
{\bf Interaction} 	        &   $2.08e+05$      &   $17$        &   $  23.04$       &   $1.91e-17$      \\ \hline
Residual                    &   $2.55e+04$      &   $48$        &                   &                   \\
\\
\emph{Freeway}              &   $sum\_sq$       &   $df$        &   $F$             &   $PR(>F)$        \\ \hline
Intercept                   &   $27174.26$      &   $1$         &   $134697.26$     &   $1.44e-59$      \\ \hline
{\bf Difficulty}		    &   $    0.28$      &   $1$         &   $     1.39$     &   $2.46e-01$      \\
{\bf Traffic}			    &   $  274.67$      &   $3$         &   $   453.84$     &   $2.73e-26$      \\
{\bf Speeds}			    &   $   18.44$      &   $1$         &   $    91.38$     &   $6.75e-11$      \\ \hline
{\bf Interaction} 		    &   $  554.30$      &   $10$        &   $   274.76$     &   $4.42e-28$      \\ \hline
Residual                    &   $    6.46$      &   $32$        &                   &                   \\
\end{tabular}
\end{table}

\begin{table}[h]
\caption{Multi-Factor ANOVA (type 3) of Default game expert zero-shot transfer vs. design factors, extended tables.}
\centering
\label{tables:anova_default_full}
\vspace{10pt}
\begin{tabular}{lcccc}
\emph{Space Invaders}       &   $sum\_sq$       &   $df$        &   $F$             &   $PR(>F)$        \\ \hline
Intercept			    	&   $3.33e+06$      &   $1$         &   $638.21$        &   $5.35e-35$      \\ \hline
{\bf Difficulty}		    &   $1.66e+06$      &   $1$         &   $319.42$        &   $1.43e-26$      \\
{\bf Invisible Invaders}	&   $1.08e+06$      &   $1$         &   $208.13$        &   $8.68e-22$      \\
{\bf Fast Bombs}		    &   $1.31e+06$      &   $1$         &   $250.95$        &   $7.92e-24$      \\
{\bf Zigzagging Bombs}		&   $3.39e+04$      &   $1$         &   $  6.51$        &   $1.32e-02$      \\
{\bf Moving Shields}	    &   $6.08e+05$      &   $1$         &   $116.63$        &   $4.67e-16$      \\ \hline
{\bf Interaction} 		    &   $1.59e+07$      &   $26$        &   $117.61$        &   $1.57e-44$      \\ \hline
Residual                    &   $3.33e+05$      &   $64$        &                   &                   \\
\\
\emph{Breakout}             &   $sum\_sq$       &   $df$        &   $F$             &   $PR(>F)$        \\ \hline
Intercept				    &   $1.36e+06$      &   $1$         &   $3436.57$       &   $2.51e-46$      \\ \hline
{\bf Difficulty}		    &   $7.23e+04$      &   $1$         &   $ 182.22$       &   $5.83e-18$      \\
{\bf Rules}			        &   $3.78e+05$      &   $2$         &   $ 476.93$       &   $2.14e-32$      \\
{\bf Extras}			    &   $2.11e+05$      &   $3$         &   $ 177.63$       &   $5.53e-26$      \\ \hline
{\bf Interaction} 	        &   $2.12e+05$      &   $17$        &   $  31.38$       &   $2.87e-20$      \\ \hline
Residual                    &   $1.90e+04$      &   $48$        &                   &                   \\
\\
\emph{Freeway}              &   $sum\_sq$       &   $df$        &   $F$             &   $PR(>F)$        \\ \hline
Intercept                   &   $1385.27$       &   $1$         &   $2775.16$       &   $1.14e-32$      \\ \hline
{\bf Difficulty}		    &   $   7.92$       &   $1$         &   $  15.87$       &   $3.66e-04$      \\
{\bf Traffic}			    &   $2282.53$       &   $3$         &   $1524.22$       &   $1.36e-34$      \\
{\bf Speeds}			    &   $ 105.63$       &   $1$         &   $ 211.61$       &   $1.17e-15$      \\ \hline
{\bf Interaction} 		    &   $ 499.23$       &   $10$        &   $ 100.01$       &   $3.14e-21$      \\ \hline
Residual                    &   $  15.97$       &   $32$        &                   &                   \\
\end{tabular}
\end{table}

\begin{figure}[h]
\begin{center}
\includegraphics[width=0.30\textwidth]{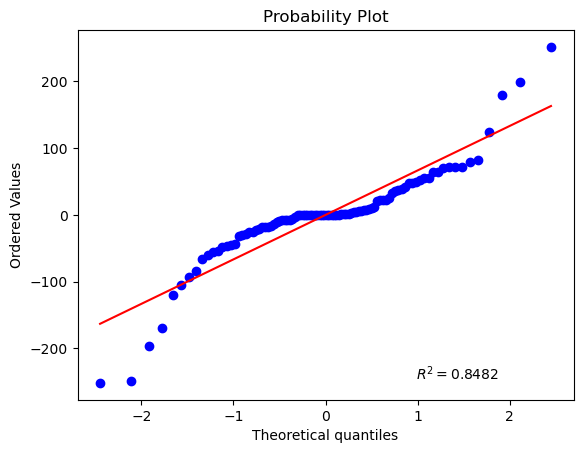}
\includegraphics[width=0.30\textwidth]{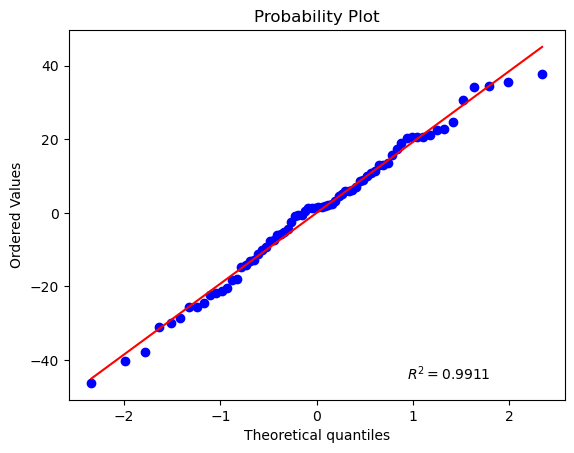}
\includegraphics[width=0.30\textwidth]{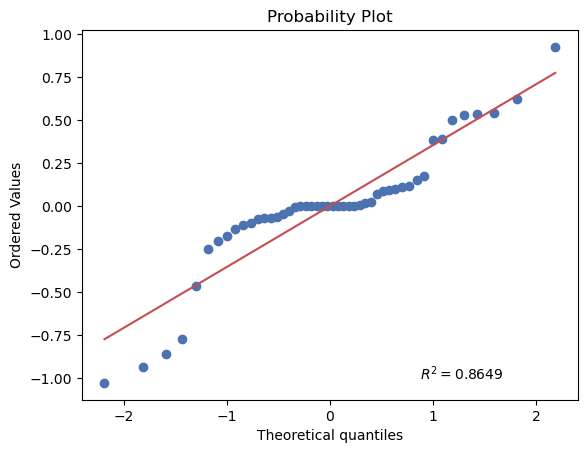}
\end{center}
\caption{Empirical distribution plots of residuals from  Multi-Factor ANOVA of Expert Scores vs. design factors.\label{figure:quantile_plots}}
\end{figure}

\subsection{Assumptions:}

\begin{enumerate}
    \item The observations in each group are independent of the observations in every other group. We trained agents independently of each other in all cases, and we selected hyper-parameters for each group (i.e. game variation) independently of others.
    \item Normality of measurement errors. For a give game, if we were to re-run the hyper-parameter grid search many times, choose the top 3 agents and average their performance, we assume that such measurements would be normally distributed. We have no reasons to believe that the distribution of the top 3 would be affected by divergence of online learning, one of the likely failure cases. We investigated empirical distribution of residuals from our analyses in Figure~\ref{figure:quantile_plots}. We observe some deviation from normality, hence it is best to only draw conclusions based on the strongest significant effects; note that deviations from normality are relevant when effect sizes are very small or when the experimental design is not balanced \citep{glass1996statistical}; in our case we use 3 samples per combination of factors (game variation), which is a balanced design. 
    \item Similar group variances of measurements. We used heteroskedasticity robust variance calculations of type ‘HC3’ \citep{davidson1993estimation} to account for observed differences in variance, which we believe are primarily due to sample size.
\end{enumerate}

\section{Appendix: Complete Results Tables}

In Table~\ref{tables:experts_sp}, Table~\ref{tables:experts_br} and Table~\ref{tables:experts_fr}  we provide raw and normalised scores for experts. In Table~\ref{tables:default_sp}, Table~\ref{tables:default_br}, Table~\ref{tables:default_fr} we list raw and normalised scores of variant-expert zero-shot transfer.

\begin{table}[h]
\centering
\caption{Space Invaders raw and variant-expert normalisation scores in various transfer learning settings and their ablations.}
\label{tables:experts_sp}
\begin{tabular}{ |C{2.3cm}||C{1.4cm}||C{1.4cm}|C{1.4cm}|C{1.4cm}||C{1.4cm}|C{1.4cm}|C{1.4cm}|} 
\hline
\multicolumn{1}{|c||}{} & \multicolumn{4}{c||}{{\bf Raw Scores}} & \multicolumn{3}{c|}{{\bf Variant-Expert Normalised Scores}} \\ \hline
{ Space~Invaders (Difficulty \& Game Mod)  } & Variant-Expert $(100\%)$  & Finetuning from Scratch & Zero-shot Expert $(0\_00)$ & Finetuned Expert $(0\_00)$ & Finetuning from Scratch $(\%)$ & Zero-shot Expert $(0\_00)$ $(\%)$& Finetuned Expert $(0\_00)$ $(\%)$ \\ \hline\hline
$0\_00$ &2167.74	&	326.68	&	2167.78	&	33090.00	&	15.07	&	100.00	&	1526.48	\\ \hline
$0\_01$ &1656.22	&	244.79	&	427.31	&	8949.71	&	14.78	&	25.80	&	540.37	\\ \hline
$0\_02$ &2250.00	&	275.85	&	1243.42	&	2820.79	&	12.26	&	55.26	&	125.37	\\ \hline
$0\_03$ &2000.00	&	235.40	&	412.60	&	2465.37	&	11.77	&	20.63	&	123.27	\\ \hline
$0\_04$ &1783.01	&	157.97	&	260.19	&	1581.55	&	8.86	&	14.59	&	88.70	\\ \hline
$0\_05$ &1170.38	&	121.72	&	89.57	&	634.09	&	10.40	&	7.65	&	54.18	\\
\hline
\end{tabular}
\end{table}

\begin{table}[h]
\centering
\caption{Breakout raw and variant-expert normalisation scores in various transfer learning settings and their ablations.}
\label{tables:experts_br}
\begin{tabular}{ |C{1.8cm}||C{1.4cm}||C{1.4cm}|C{1.4cm}|C{1.4cm}||C{1.4cm}|C{1.4cm}|C{1.4cm}|} 
\hline
\multicolumn{1}{|c||}{} & \multicolumn{4}{c||}{{\bf Raw Scores}} & \multicolumn{3}{c|}{{\bf Variant-Expert Normalised Scores}} \\ \hline
{ Breakout (Difficulty \& Game Mode)  } & Variant-Expert $(100\%)$  & Finetuning from Scratch & Zero-shot Expert $(0\_00)$ & Finetuned Expert $(0\_00)$ & Finetuning from Scratch $(\%)$ & Zero-shot Expert $(0\_00)$ $(\%)$& Finetuned Expert $(0\_00)$ $(\%)$ \\ \hline\hline 
$0\_00$ &264.60	&	6.69	&	264.60	&	342.21	&	2.53	&	100.00	&	129.33	\\ \hline
$0\_01$ &420.10	&	5.25	&	212.41	&	412.43	&	1.25	&	50.56	&	98.17	\\ \hline
$0\_02$ &298.60	&	6.51	&	201.41	&	325.49	&	2.18	&	67.45	&	109.01	\\ \hline
$0\_03$ &247.00	&	6.05	&	22.02	&	69.99	&	2.45	&	8.92	&	28.33	\\ \hline
$0\_04$ &71.90	&	82.43	&	116.85	&	111.95	&	114.65	&	162.51	&	155.70	\\ \hline
$0\_05$ &55.50	&	70.97	&	106.79	&	101.32	&	127.87	&	192.42	&	182.56	\\
\hline
\end{tabular}
\end{table}

\begin{table}[h]
\centering
\caption{Freeway raw and variant-expert normalisation scores in various transfer learning settings and their ablations.}
\label{tables:experts_fr}
\begin{tabular}{ |C{1.8cm}||C{1.4cm}||C{1.4cm}|C{1.4cm}|C{1.4cm}||C{1.4cm}|C{1.4cm}|C{1.4cm}|} 
\hline
\multicolumn{1}{|c||}{} & \multicolumn{4}{c||}{{\bf Raw Scores}} & \multicolumn{3}{c|}{{\bf Variant-Expert Normalised Scores}} \\ \hline
{ Freeway (Difficulty \& Game Mode)  } & Variant-Expert $(100\%)$  & Finetuning from Scratch & Zero-shot Expert $(0\_00)$ & Finetuned Expert $(0\_00)$ & Finetuning from Scratch $(\%)$ & Zero-shot Expert $(0\_00)$ $(\%)$& Finetuned Expert $(0\_00)$ $(\%)$ \\ \hline\hline 
$0\_00$ & 33.10	&	23.31	&	33.14	&	33.43	&	70.43	&	100.12	&	100.99	\\ \hline
$0\_01$ & 25.83	&	5.94	&	0.37	&	23.55	&	22.99	&	1.43	&	91.18	\\ \hline
$0\_02$ & 27.13	&	4.98	&	0.10	&	11.14	&	18.35	&	0.38	&	41.07	\\ \hline
$0\_03$ & 32.97	&	11.06	&	0.24	&	24.90	&	33.55	&	0.73	&	75.52	\\ \hline
$0\_04$ & 33.30	&	21.86	&	18.30	&	24.55	&	65.64	&	54.94	&	73.73	\\ \hline
$0\_05$ & 25.65	&	7.96	&	1.74	&	9.94	&	31.03	&	6.80	&	38.74	\\ \hline
$0\_06$ & 25.66	&	7.98	&	1.53	&	10.10	&	31.10	&	5.98	&	39.38	\\ \hline
$0\_07$ & 28.74	&	13.23	&	1.19	&	16.49	&	46.03	&	4.13	&	57.38	\\ \hline\hline
$1\_00$ & 33.08	&	16.94	&	33.06	&	33.31	&	51.20	&	99.93	&	100.68	\\ \hline
$1\_01$ & 26.87	&	5.31	&	0.23	&	23.86	&	19.75	&	0.86	&	88.79	\\ \hline
$1\_02$ & 27.34	&	0.74	&	0.00	&	0.00	&	2.69	&	0.01	&	0.00	\\ \hline
$1\_03$ & 32.88	&	8.16	&	0.01	&	23.86	&	24.83	&	0.02	&	72.56	\\ \hline
$1\_04$ & 33.39	&	17.98	&	13.16	&	23.37	&	53.85	&	39.41	&	69.98	\\ \hline
$1\_05$ & 25.85	&	4.92	&	1.01	&	7.94	&	19.03	&	3.90	&	30.71	\\ \hline
$1\_06$ & 25.66	&	4.67	&	0.74	&	7.06	&	18.21	&	2.89	&	27.52	\\ \hline
$1\_07$ & 28.81	&	7.67	&	0.42	&	14.24	&	26.63	&	1.48	&	49.44	\\ \hline
\end{tabular}
\end{table}

\begin{table}[h]
\centering
\caption{Space Invaders zero-shot transfer matrix between variant-experts trained from scratch. Top: raw scores, Bottom: Variant-Expert normalised scores (percentages).}
\label{tables:default_sp}
\begin{adjustbox}{angle=90}
\resizebox{.98\textheight}{!}{%
}
\end{adjustbox}
\end{table}

\begin{table}[h]
\centering
\caption{Breakout zero-shot transfer matrix between variant-experts trained from scratch. Top: raw scores, Bottom: Variant-Expert normalised scores (percentages).}
\label{tables:default_br}
\begin{adjustbox}{angle=90}
\resizebox{.98\textheight}{!}{%
}
\end{adjustbox}
\end{table}

\begin{table}[h]
\centering
\caption{Freeway zero-shot transfer matrix between variant-experts trained from scratch. Top: raw scores, Bottom: Variant-Expert normalised scores (percentages).}
\label{tables:default_fr}
\resizebox{.99\textwidth}{!}{%
}
\end{table}

\end{document}